\documentclass{ifacconf}

\usepackage{graphicx}
\usepackage{natbib}
\usepackage{amsmath,amsfonts,bm,amssymb}
\usepackage{xcolor}
\usepackage{booktabs,multirow}
\usepackage{empheq}
\usepackage{xstring}
\usepackage[hyphens]{url}

\def\eqref#1{(\ref{#1})}
\def\1{\bm{1}}
\newcommand{\train}{\mathcal{D}}

\def\rs{{\textnormal{s}}}

\def\rve{{\mathbf{e}}}

\def\rvu{{\mathbf{i}}}

\def\rvq{{\mathbf{q}}}
\def\rvr{{\mathbf{r}}}
\def\rvs{{\mathbf{s}}}

\def\rvu{{\mathbf{u}}}
\def\rvv{{\mathbf{v}}}
\def\rvw{{\mathbf{w}}}
\def\rvx{{\mathbf{x}}}
\def\rvy{{\mathbf{y}}}
\def\rvz{{\mathbf{z}}}

\def\ve{{\bm{e}}}

\def\vs{{\bm{s}}}

\def\vu{{\bm{u}}}
\def\vv{{\bm{v}}}

\def\vx{{\bm{x}}}
\def\vy{{\bm{y}}}
\def\vz{{\bm{z}}}

\DeclareMathAlphabet{\mathsfit}{\encodingdefault}{\sfdefault}{m}{sl}
\SetMathAlphabet{\mathsfit}{bold}{\encodingdefault}{\sfdefault}{bx}{n}

\def\sR{{\mathbb{R}}}
\def\sS{{\mathbb{S}}}

\def\sU{{\mathbb{U}}}
\def\sV{{\mathbb{V}}}
\def\sW{{\mathbb{W}}}

\def\sY{{\mathbb{Y}}}
\def\sZ{{\mathbb{Z}}}

\newcommand{\E}{\mathbb{E}}

\newcommand{\KL}{D_{\mathrm{KL}}}

\usepackage{amssymb}
\usepackage{graphicx}
\usepackage{xcolor}         % colors
\usepackage{booktabs,multirow}

\newcommand{\reals}{\mathbb{R}}

\newcommand{\bP}{\mathbf{P}}
\newcommand{\F}{\mathbf{F}}

\newtheorem{Assumption}{Assumption}[section]
\newtheorem{lemma}{Lemma}[section]
\newtheorem{corollary}{Corollary}[section]

\newtheorem{Remark}{Remark}[section]
\newtheorem{Definition}{Definition}[section]
\newtheorem{Theorem}{Theorem}[section]

\usepackage{empheq}

\newcommand{\class}{\(\mathcal{SI}\)}
\newcommand{\classOld}{\(\mathcal{S}\)}

\newcommand{\deividas}[1]{\textcolor{black}{ #1}} %\marginpar{\textcolor{blue}{Deividas}}

\usepackage{xstring}
\newcommand{\mref}[1]
{%
	\IfEqCase{#1}{%
		{lemma:full_gen_sys}{A.7}% error system definition
		{lemma:stationary_x_v}{A.4}%replaces (Theorem 5.1) 
		{lemma:classExamples}{A.6}%replaces (Theorem 5.2) 
		{lemma:series}{A.8}%
		{lemma:Cross-EntropyLipschitz}{A.10}%
		{eq:numexGen}{213}%
		{sec:exampleAppendix}{B}%
		{sec:proofsResults}{A.5}
	}[\ref{#1}]%
}

\begin{document}

\begin{frontmatter}

\title{ From PAC bounds on the generalization error to PAC bounds on the parameter estimation error.}

\author[ULille]{Mih\'aly Petreczky}
\author[AAU]{John Leth}
\author[SZTAKIELTE]{Emma Luk\'acs}
\author[SZTAKI]{B\'alint Dar\'oczy}
\address[ULille]{Univ. Lille, CNRS, Centrale Lille, UMR 9189 CRIStAL, F-59000 Lille, France (e-mail: mihaly.petreczky@centralelille.fr).}
\address[AAU]{Department of Electronic Systems, Aalborg University, Denmark (e-mail: jjl@es.aau.dk).}
\address[SZTAKIELTE]{Institute for Computer Science and Control, HUN-REN SZTAKI, Budapest, Hungary \& Institute of Mathematics, ELTE, Budapest, Hungary (e-mail: emma.lukacs@sztaki.hu).}
\address[SZTAKI]{Institute for Computer Science and Control, HUN-REN SZTAKI, Budapest, Hungary (e-mail: balint.daroczy@sztaki.hu).}
%\address[ELTE]{Institute of Mathematics, Eötvös Loránd University, ELTE, Budapest, Hungary}

\begin{abstract}
Probably Approximately Correct (PAC) bounds on the generalization error, the difference between the true prediction error and the empirical prediction error, are an established tool in statistical machine learning, and there have been several extensions of these bounds to dynamical systems. In parallel, in recent years, there has been growing interest in finite-sample bounds for parameter estimation error in system identification. The latter bounds tend to be algorithm-specific, relying on reformulating the identification algorithm as a least-squares regression problem. This is in contrast to classical PAC bounds, which are algorithm-agnostic and can be applied to any identification algorithm based on minimizing a cost function involving the empirical loss.
In this paper, we show how oracle bounds derived from PAC bounds can be used to obtain finite-sample bounds for dynamical systems. As an illustration of this approach, we present finite-sample parameter-estimation bounds for linear and bilinear systems with bounded inputs and noise. These bounds apply to identification algorithms based on minimizing the empirical loss function.
%The key ingredient is to establish an oracle inequality, i.e., an inequality between the true prediction error of the learned model and the minimum achievable true prediction error, and then use either a quadratic-growth condition on the true loss or system-theoretic properties of the loss function to derive bounds on the parameter estimation error. 
\end{abstract}

\end{frontmatter}

\section{Introduction}

The problem of learning from single observed trajectories raises two interconnected questions about the performance of the learned model: prediction quality for future observations and parameter estimation error (recovery of the underlying system).
The statistical learning literature typically focuses on prediction, while the system identification literature emphasizes parameter estimation.
This distinction arises because machine learning models are usually applied to inputs sampled from distributions similar to the training data, whereas system identification aims to produce models for decision-making (controller synthesis, fault detection, etc.) under distribution shift.

In statistical learning, predictive performance is typically characterized by finite-sample \textbf{(1)} \emph{generalization bound} which upper-bounds the gap between the \emph{true error} (expected prediction error under the data distribution) and the \emph{empirical error} (average prediction error on training data), and \textbf{(2)} \emph{oracle (excess risk) bound} which upper-bounds the gap between the learned model's true error and the best achievable true error. Both bounds depend on training set size and ideally decrease as this size increases.

System identification focuses on parameter estimation consistency: as data grows, the estimated parameter converges to the minimizer of the true error. Under identifiability and persistence of excitation, this minimizer is unique. If it coincides with the true system parameter, the estimated parameter converges to the true parameter. %Beyond asymptotic consistency, the literature also establishes that the normalized parameter estimation error is asymptotically normal.

However, these asymptotic results provide limited insight for finite samples, which is crucial when combining system identification with control design (adaptive control, data-driven control, reinforcement learning). This motivates recent interest in finite-sample bounds on the parameter estimation error (see \cite{Ziemann1}).
Existing results tend to be geared towards specific algorithms based on linear regression and Ho-Kalman or subspace methods, and do not apply to prediction error minimization (PEM) for general state-space representations. In contrast, generalization and oracle bounds in the machine learning literature typically apply to PEM (empirical risk minimization in ML terminology). 

%To fill this gap, we aim to transpose PAC bounds on generalization error/excess risk to
%parameter estimation bounds. The challenge is that 
%a-priori small generalization error does not guarantee small estimation error.
% a key question arises: how can we convert bounds on generalization error into bounds on parameter estimation error?

\textbf{Contribution}
In this paper we connect these two frameworks for several families of stable dynamical systems under bounded stochastic inputs and noise. First, we derive a PAC bound on the generalization error for learning stable non-linear dynamical systems from a single time-series. We then derive oracle inequalities from this generalization bound for prediction error minimization (PEM) algorithms and use them to obtain finite-sample parameter estimation bounds. For linear and bilinear systems, we exploit Markov parameters and the $H_2$ norm to bound the distance between learned and true systems. The squared estimation error decays at rate $O(\log N/\sqrt N)$, which is slower than some existing results and applies only to bounded data. However, our unified approach covers a large class of models and algorithms, revealing the systematic relationship between prediction and parameter estimation error.

%under a quadratic growth condition and the excitation conditions yield $H_2$ distance bounds between the Markov parameters of the learned and the true system. 

%We give proof sketches, full proofs \cite{arxiv}.
Majority of the proofs are in the Appendix. 

\textbf{Related work}
  There is a rich literature on finite-sample bounds for parameter estimation in LTI systems, e.g., \cite{simchowitz2019learning,oymak2021revisiting,Pappas1,Ziemann1,Ziemann2}, and for switched systems \cite{racz2025finite} and bilinear systems \cite{pmlr-v283-sattar25a}. Most existing results address Ho-Kalman realization algorithms or subspace methods and do not handle general partially-observed state-space models via prediction error minimization (PEM).
 \par 
  Oracle inequalities are standard in statistical learning with i.i.d. data \cite{shalev2014understanding} and extend to autoregressive models \cite{alquier2013prediction}. \citet{eringis2023pacbayesRNN} derived PAC-Bayesian bounds on generalization error for RNNs by exploiting stability, but without oracle inequalities or parameter estimation bounds. Related work in \cite{Nips2026} applies similar techniques to LTI systems; the present paper extends this to general nonlinear systems while simplifying the proof structure.
  The advantage of \cite{eringisRenyi,Nips2026} over our approach is the ability to handle unbounded data, whereas we require boundedness. However, our unified treatment covers a large class of nonlinear models and reveals the systematic connection between prediction error and parameter estimation error.

%We seek to connect these statistical learning techniques of measuring performance with system identification.
% The standard i.i.d. condition in statistical learning with a few recent exceptions prevented the adaptation of the Probably Approximately Correct (PAC) framework for single trajectory problems. \citet{alquier2013prediction} established oracle inequalities for time series forecasting while assuming bounded weakly dependent processes.. In parallel, in \citet{Simchowitz_Mania_Tu_Jordan_Recht_2018} the authors analyzed the error of identification of linear systems from a single trajectory. These complementary results address two distinct although aligned problems. How well the learned model predicts future observations, generalization, and how the learned parameters differ from the underlying system's parametrization, estimation.  

%in In this paper we only show the idea of the proofs, for more details see \cite{arxiv}. 

\textbf{Notation}.
Note that unless otherwise defined, this paper will follow the notation defined by \citet{goodfellow2016deep}, i.e.  $\vx$ is a vector, and $\rvx$ is a random vector.
Let $\F$ denote a $\sigma$-algebra on the set $\Omega$ and $\bP$ be a probability measure on $\F$. Unless otherwise stated all probabilistic considerations will be with respect to the probability space $(\Omega,\F,\bP)$, and we let $\E(\rvz)$ denote expectation of the stochastic variable \deividas{$\rvz:\Omega\rightarrow\reals^{n_\rvz}$. We shall denote the realisation of a stochastic variable $\rvz$ as $\vz=\rvz(\omega)$, with $\omega\in\Omega$.}
Each Euclidean space is associated with the topology generated by the 2-norm $\|\cdot\|_2$, and the Borel $\sigma$-algebra generated by the open sets. The induced matrix 2-norm is also denoted $\|\cdot\|_2$. 
We use $\triangleq$ to denote ''defined by''.%, and $\aseq$ to denote that the equality holds almost surely with respect to some underlying probability measure. 

%\section{Problem Formulation}
\section{Problem Formulation}
% \section{PAC-Bayesian Setting and Main Result}

\label{sec:prelim}
%In this paper we will consider time-series supervised learning problem. The goal will be to find a model which predicts the best predictive power.
% To do this we assume we have only one sequence of training data.
%Below  we state the system identification/learning problem.
 Let us fix \emph{bounded} stochastic processes $\rvy(t)\in \sY\subset \reals^{n_y},\rvu(t)\in \sU \subset \reals^{n_x}$ that share the time-axis $t\in\sZ$, i.e. $\rvy(t),\rvu(t)$ are random vectors on $(\Omega,\F,\bP)$; $\rvy$ and $\rvu$ represent outputs and inputs respectively.
 We set $\rvw(t)=\begin{bmatrix} \rvy^T(t)& \rvu^T(t) \end{bmatrix}^T$ and $\sW=\sY \times \sU$.  We assume that $\rvy$ is generated by
 a dynamical system of the form
 \begin{equation}\label{eq:gen}
    \rvs(t+1)=f_g(\rvs(t),\rve_g(t)), \quad  \rvw(t) =g_g(\rvs(t),\rve_g(t)) 
\end{equation}
where $\rve_g$ is a bounded i.i.d. process in some set $\sV_g \subseteq \reals^{n_v}$ and we assume that the solution $\rvs(t)$ is well-defined for all $t \in \mathbb{Z}$ and lives in a state-space $\sS_g \subseteq \reals^{n_s}$. We will make additional
assumptions later on.

The goal of a predictor $h\in\mathcal{H}$ is to estimate the current output based on past outputs and inputs.
%$\vw(t)=\rvw(t)(\omega)$ up to initial time.
 Formally, a predictor $h$ is a 
function $h:\bigcup_{k=1}^{\infty} \sU^{k-1} \times \sY^{k-1} \rightarrow \sY$ 
generating the prediction $\hat{\rvy}(t)(\omega)=h(\{\rvu(s)(\omega)\}_{s=0}^t,\{\rvy(s)(\omega)\}_{s=0}^{t-1})$.
We consider predictors realized by parametrizations of dynamical systems; we
assume that there exists a parameter ser $\Theta\subseteq \reals^{n_\theta}$, a compact set $\sS \subseteq \mathbb{R}^n$,
a function $\Pi: \Theta \rightarrow \mathcal{H}$, and continuous functions
\begin{align*}
    \hat{f}:\sS \times \sW \times \Theta \rightarrow \sS, \quad
    \hat{g}: \sS \times \sU \times \Theta \rightarrow \sY, \quad 
    \hat{\vs}_s:\Theta \rightarrow \sS
\end{align*}
such that  for every predictor $h_{\theta}=\Pi(\theta)$, every $\mathbb{Z} \ni t \ge t_0$
\begin{color}{black}
\begin{equation}\label{eq:dyn1}
\mathcal{S}_{\theta}\left\{
\begin{aligned}
  & \hat{\rvs}(t+1 \!\!\mid\!\! t_0)=\hat{f}\bigl(\hat{\rvs}(t \!\!\mid\!\! t_0),\rvy(t),\rvu(t),\theta\bigr), \\
  & \hat{\rvy}_{\theta}(t \!\!\mid\!\! t_0)=\hat{g}\bigl(\hat{\rvs}(t\!\! \mid\!\! t_0),\rvu(t),\theta\bigr), ~~ \hat{\rvs}(t_0 \!\!\mid\!\! t_0)=\hat{\vs}_s(\theta),\\
  & h_{\theta}\bigl(\{\rvu(s)\}_{s=t_0}^t,\{\rvy(s)\}_{s=t_0}^{t-1}\bigr)=\hat{\rvy}_{\theta}(t \!\!\mid\!\! t_0)
\end{aligned}
\right.
\end{equation}
\end{color}
%where $\hat{\rvs}(t)\in \sS$ is the hidden state. 
Moreover, we assume that \eqref{eq:dyn1} has a unique stationary solution 
$\hat{\rvs}(t)$ defined on all $t \in \mathbb{Z}$ satisfying 
\[ 
    \hat{\rvs}(t+1)=\hat{f}(\hat{\rvs}(t),\rvy(t),\rvu(t),\theta), \quad \hat{\rvy}_{\theta}(t)=\hat{g}(\hat{\rvs}(t),\rvu(t),\theta)
\]
such that $\hat{\rvs}_{\theta}$, $\hat{\rvy}_{\theta}$ are stationary processes, and they satisfy
\begin{equation}
  \label{limits_true_loss}
  \begin{split}
      & \hat{\rvs}(t)=\lim_{t_0 \to -\infty} \hat{\rvs}(t \!\!\mid\!\! t_0),  ~
      \hat{\rvy}_{\theta}(t)=\lim_{t_0 \to -\infty} \hat{\rvy}_{\theta}(t \!\!\mid\!\! t_0) \\
%&\lim_{t\to\infty} \mathbb{E}\!\left[
%  \left\|\hat{y}_\theta(t)-\hat{y}_\theta(t\mid t_0)\right\|_2
%\right] = 0, \forall t_0
%      & \lim_{t \rightarrow \infty} (\hat{\rvy}_{\theta}-\hat{\rvy}_{\theta}(t \mid t_0))=0 \\
%      & \E[\hat{\rvy}_{\theta}(t)]=\lim_{t_0 \to -\infty} \E[\hat{\rvy}_{\theta}(t \mid t_0)]=\lim_{t \rightarrow \infty} \E[\hat{\rvy}_{\theta}(t)-\hat{\rvy}_{\theta}(t \mid 0)]
  \end{split}
\end{equation}      
  and the limits of the random variables are understood in the mean sense. 
We will add additional stability assumptions later on which guarantee that \eqref{limits_true_loss} holds.
%existence of $\hat{\rvs}$ and the limits mentioned above. 
Intuitively, $\hat{\rvy}_{\theta}(t)$ is the prediction generated by the predictor if it has been 
active for an infinite amount of time.

%In order to simplify notation, we 
%\begin{equation}
%\label{as:parametrisation:eq1}
%\begin{split}
%	\hat{\rvy}_{\theta}(t|0) &\triangleq \textcolor{black}{\rvy_{S_{\theta}}(\hat{\rvs}_{s}(\theta),0,\rvw;t)=h_{\theta}(\{\rvw(s)\}_{s=0}^t)}, \\ 
%\end{split}
%\end{equation}
%Mi.e., $\hat{\rvy}_{\theta}(t|0)$ is just the prediction at time $t$ generated by $h_{\theta}$, and 
%$\hat{\rvy}_{\theta}(t)$ is the output process
%	$\rvy_{S_\theta,\rvw}(t)$ of the \class{} system  $S_\theta$ associated  with $\rvw$.

%Note that recurrent neural networks (RNNs%)
%represent a special case of the dynamical systems of the form of \eqref{eq:dyn1}. 
%Under suitable assumptions, which will be discussed in Section \ref{sec:ProblemForm}, the generalisation gap is bounded for non i.i.d. data, and predictors realised by dynamical systems. 
The desired model should be learnt from a data set $\train=\{\rvy(s)(\omega),\rvw(s)(\omega)\}_{s=0}^{N-1}$ formed by a single sample of random variables $\{\rvy(s),\rvw(s)\}_{s=0}^{N-1}$ We choose a Lipschitz loss function $\ell: \sY \times \sY \to \sR_+$, and 
we measure the quality of the predictor $\theta$ by  the \emph{true error}
\begin{equation}
    \label{gen:losses}
    \mathcal{L}(\theta) \triangleq \E[\ell(\rvy(t),\hat{\rvy}_{\theta}(t))]
\end{equation}
%For now, we assume that the limit  defining the generalisation loss $\mathcal{L}(h)$ exists, later we define $\mathcal{H}$ so that this  assumption holds. 
When learning, we would like to choose a predictor $\hat{\theta}_N$ which minimizes $\mathcal{L}(\theta)$. Since
$\mathcal{L}(\theta)$ is unknown, we instead minimize the 
\emph{empirical  error (loss)} %$\hat{\mathcal{L}}_N(h)$ and $\mathcal{L}(h)$:
% \begin{Definition}
\begin{equation}
    \label{prelim:losses}
    \hat{\mathcal{L}}_N(\theta)\triangleq\frac{1}{N}\sum_{t=0}^{N-1}\ell(\rvy(t),\hat{\rvy}_{\theta}(t \!\!\mid\!\! 0))
\end{equation} 
In system identification the empirical loss is  often defined
as $V_N(\theta)=\frac{1}{N}\sum_{t=0}^{N-1}\ell(\rvy(t),\hat{\rvy}_{\theta}(t)$, i.e., $\hat{\rvy}_{\theta}(t)$  is used instead of $\hat{\rvy}_{\theta}(t \mid 0)$
in \eqref{prelim:losses}, by assuming that the predictors are already in their steady state regime when they are applied to the data.
This is a reasonable when analyzing asymptotic behavior of learning algorithms (i.e., $N \rightarrow \infty$), but not in the finite-sample setting. By \cite{eringis2023pacbayesRNN}
$\lim_{N \rightarrow \infty} \left(\hat{\mathcal{L}}_N(\theta)-V_N(\theta)\right) = 0$ in the mean sense.
A PAC bound is a probabilistic bound on the generalisation gap
$\mathcal{L}(h)-\hat{\mathcal{L}}_N(h)$, i.e.,  an inequality of the form 
%since bounding the gap will provide immediate bounds on $\mathcal{L}(h)$
%That is, in this paper we shall prove a \emph{particular PAC inequality}.
%Informally, let $\mathcal{H}$ be a family
%of predictors which can be realised by a 
%dynamical system of the form \eqref{eq:dyn1}. for any $\delta \in (0,1/2)$, the following holds with probability at least $1-2\delta$
    \begin{equation} \label{eq:pac:gen}
        \bP(\forall \theta \in \Theta: |\mathcal{L} (\theta)- \hat{\mathcal{L}}_N(\theta)| \le 
        \varepsilon_N(\delta)) \ge 1-2\delta
    \end{equation}
     for all $\delta \in (0,1/2)$, where $\varepsilon_N(\delta):=2\left(\ln\frac{2}{\delta}+C_{\mathrm{pac}}\right)\frac{\ln N}{\sqrt N}$. 
  %Such a bounds
  % relates true loss with the empirical error
  %and can be used for analyzing the outcome of 
  We can use this bound for analyzing learning algorithms based
on minimizing the empirical error:
\[\hat{\theta}_N = \arg\min_{\theta \in \Theta} \hat{\mathcal{L}}_N(\theta), \]
%Intuitively,  \eqref{eq:pac:gen} implies that for large $N$ minimizing the empirical error is similar to minimizing the
%true one.
%$\mathcal{L} (\hat{\theta}_N)$ and $\hat{\mathcal{L}}_N(\hat{\theta}_N)$ are close, so minimizing
%the latter will make the former minimal. 
as with standard arguments \eqref{eq:pac:gen} leads to \emph{oracle inequality}
\begin{equation}
\label{eq:oracle}
\bP(\mathcal{L}(\hat{\theta}_N) \le \inf_{\theta \in \Theta} \mathcal{L}(\theta) + 2\varepsilon_N(\delta)) \ge 1-2\delta 
\end{equation}
which implies that the true error of $\hat{\theta}_N$ is close to the minimal one. 
%formalizes the intuition that the true loss of the empirical loss minimizer is 
%get close to the minimizer of the true loss as $N \rightarrow +\infty$.
However, in system identification, 
we would like to show that $\hat{\theta}_N$ is close to the minimizer $\theta_{true}$ of the true error,
under identifiability and persistence of excitation assumptions, for a suitable distance $d$ on models. 
%\mathrm{argmin}_{\theta \in \Theta} \mathcal{L}(\theta)$ 
%What we need is to 1) relate the achieved true losss
%$\mathcal{L}(\hat{h}_N)$ with the best achievable true loss 
%$\inf_{h \in \mathcal{H}} \mathcal{L}(h)$, 2)
%if $h_{true}$ is the minimizer of 
%$\inf_{h \in \mathcal{H}} \mathcal{L}(h)$, 
%then we would like a
%we would like to bound the model estimation error $d(h,h_{true})$ where $h_{true}=\mathrm{argmin}_{h \in \mathcal{H}} \mathcal{L}(h)$
%is the minimizer of the true loss and $d$
%for a suitable metric on the space of model classes. 

In this paper, we derive a version of the PAC bound \eqref{eq:pac:gen} and of its oracle version \eqref{eq:oracle}, and we use
these to show a bound on the parameter estimation error
\begin{equation}
  \label{eq:param_error}
  \begin{split}
  \mathbb P\!\left(
d(\widehat\theta_N,\theta_{\mathrm{true}})^2
\le C_{\mathrm{param}}\,\varepsilon_N(\delta)
\right)\ge 1-2\delta,
%\bP(d(\hat{\theta}_N,\theta_{true}) \le (1+\frac{\ln(\frac{2}{\delta}))C_{param}}{\sqrt{N}}) \ge 1-\delta 
\end{split}
\end{equation}
therefore %the squared estimation error is bounded at rate $O(\log N/\sqrt N)$, and
the estimation error itself at rate $O(\sqrt{\log N}/N^{1/4})$.
Note that \eqref{eq:param_error} and \eqref{eq:oracle} imply asymptotic consistency of the learning based on minimizing the empirical loss, but they also
allow to estimate the estimation error for a certain number of data points. %$N$. 
In particular, it can be used to estimate the minimal
number $N_{\epsilon}$ of the data points which guarantees that the estimation error is smaller than some threshold $\epsilon > 0$.

\section{Class of Dynamical Systems}\label{sec:class}
%In this paperconsider predictors realised by a special class of dynamical systems. Furthermore, later we assume that the data is an output of such a system. In this section, we
%Mdefine this class of systems.
We restrict the class of dynamical systems \eqref{eq:gen}-\eqref{eq:dyn1} realizing predictors to be of class \class{} defined below. 
%\deividas{To this end, let us consider a deterministic version the dynamical system in \eqref{eq:dyn1}:}
%\eqref{eq:generalNLsys}. 
%(generalisation of recurrent neural networks)
\begin{subequations}\label{eq:generalNLsys}
%\Sigma empheqlbrac
%\begin{cases}
\begin{empheq}[left=S: \empheqlbrace]{align}   \vs(t+1)&=f(\vs(t),\vv(t)),\label{eq:generalNLsys_f}\\
    \vy(t)&=g(\vs(t),\vv(t)), \label{eq:generalNLsys_h}
\end{empheq}
\end{subequations}
where $\vv: \mathbb{T} \rightarrow \sV$ is an arbitrary input trajectory \deividas{(e.g., a realization $\vx(t)=\rvw(t)(\omega)$ as in \eqref{eq:dyn1})},  $\vs: \mathbb{T} \rightarrow \sS$ is the state trajectory,  
 $\vy: \mathbb{T} \rightarrow \sY$ is the output  trajectory, and $\mathbb{T} \subseteq \sZ$ is an interval in $\sZ$
 which has no upper bound.
 That is,  either $\mathbb{T}$ is the whole $\sZ$ or
 $\mathbb{T}$ is an interval
 $[t_0, +\infty) \cap \mathbb{Z}$
 for some $t_0$.
 Moreover, $\sS\subseteq\sR^n$ is a compact set, and the sets 
  $\sV\subset\sR^m$, and $\sY\subset\sR^p$ are bounded, 
  \textcolor{black}{$\sS,\sV,\sY$ all contain the zero vectors and 
  $f(0,0)=0$, $g(0,0)=0$.}
% \par
%The state trajectory $\vs$ is also referred to as a \emph{solution}. 
If $\mathbb{T}=[t_0, +\infty) \cap \mathbb{Z}$, then 
$\vs$ is uniquely determined by the initial state 
$\vs(t_0)=\vs_0$ and the input $\vv$, and 
% A solution to \eqref{eq:generalNLsys_f} is sometimes written as 
we write $\vs(t)=\vs_{S}(\vs_0,t_0, \vv;t)$ to emphasise the dependence on initial conditions $\vs(t_0)=\vs_0$, and input $\vv$.
%\mathbb{T}\to\sV$, and the 
%system $S$. 
Moreover, if $\vv$ is defined on the whole time axis $\mathbb{Z}$,
%instead of
%$\mathbb{T}$, 
then $\vs_{S}(\vs_0,t_0, \vv;t)$
is understood as the solution %$\vs(\vs_0,t_0,\vv|_{\mathbb{T}};t)$ 
corresponding to the restriction of 
$\vv$ to $\mathbb{T}$.
%When $S$ is clear from the context, 
%we use $\vs(\vs_0,t_0, \vv;t)$ instead
%of $\vs_{S}(\vs_0,t_0, \vv;t)$.
%\par
Likewise, we will use 
$\vy_{S}(\vs_0,t_0,\vv;t)$ to denote
the output $\vy(t)$ which corresponds 
to the state trajectory $\vs(t)=\vs_{S}(\vs_0,t_0, \vv;t)$, i.e., $\vy_S(\vs_0,t_0,\vv;t)=g(\vs_{S}(\vs_0,t_0, \vv;t),\vv(t))$. 
%As before, we drop the subscript $S$ 
%from $\vy_{S}(\vs_0,t_0, \vv;t)$, 
%if it  is clear from the context.
We \emph{identify systems of the form (\ref{eq:generalNLsys})
with the \begin{color}{black} tuple \end{color}$S=(\sS,\sY,\sV,f,g)$}.

\begin{Definition}[Class \class{} system] \label{def:classDef}
We say that %the system from 
\eqref{eq:generalNLsys} is a class \class{} system, %or the pair $(f,g)$ is of class \class{}, 
with positive constants $C=C(S)\geq 1,\tau=\tau(S)\in[0,1),L=L(S),L_{g,1}=L_{g,1}(S)$, and $L_{g,2}=L_{g,2}(S)0$ if the following holds:
%\begin{itemize}

\textbf{(UEC)} $S$ is \emph{uniformly exponentially convergent with constants $C$ and $\tau$}
   for each bounded input $\vv:\sZ \rightarrow \sV$, there exists a unique state trajectory $\vs=\vs_{S,\vv}: \sZ \rightarrow \sS$ of $S$,
        and for any initial state $\vs_0 \in \mathbb{S}:$
        %\item There exist $C\geq 1$ and $\tau\in[0,1)$ s.t.,
        \begin{equation*}
            \|\vs_S(\vs_0,t_0,\vv;t)-\vs_{S,\vv}(t)\|_2\leq C\tau^{t-t_0}\|\vs_0-\vs_{S,\vv}(t_0)\|_2
        \end{equation*}
   We refer to $\rvs_{S,\rvv}$     
as the \emph{steady-state state trajectory}
of $S$ associated with $\rvv$.
We call
$\rvy_{S,\rvv}:\sZ \ni t \mapsto g(\rvs_{S,\rvv}(t),\rvv(t))$
the \emph{steady-state output trajectory} associated with $\rvv$.

\begin{color}{black}
\textbf{(UEIIS)} $S$ is \emph{uniformly exponentially incrementally input-to-state stable with constants $C,\tau$}, i.e.,
  for every $\vs_{0,1},\vs_{0,2} \in \sS$  and every two bounded inputs $\vv_1,\vv_2:\sZ \rightarrow \sV$,
  \[ 
    \begin{split}
     & \|\vs_{\mathcal{S}}(\vs_{0,1},t_0,\vv_1;t)-\vs_{\mathcal{S}}(\vs_{0,2},t_0,\vv_2;t)\|_2\leq \\ 
     & C\tau^{t-t_0}\|\vs_{0,1}-\vs_{0,2}\|_2  
      + L\sum_{k=1}^{t-t_0} \tau^{k-1} \|\vv_1(t-k)-\vv_2(t-k)\|_2
  \end{split}
  \]
\end{color}
%\item  
\textbf{(Lipschitz output)}
    The output function $g$ %from \eqref{eq:generalNLsys_h} 
    has Lipschitz  constants $L_{g,1},L_{g,2}>0$, i.e.,
    %\begin{align}
     \(   \|g(\xi_1,v_1)-g(\xi_2,v_2)\|_2\leq L_{g,1}\|\xi_1-\xi_2\|_2+L_{g,2}\|v_1-v_2\|_2 \), 
     for all $\xi_1,\xi_2 \in \sS$, $v_1,v_2 \in \sV$.
    %\end{align}
%\end{itemize}
% We shall denote by $G($
\end{Definition}

%UEC implies robustness with respect to perturbations in the initial state. 
%Exponential robustness in inputs implies that state is robust w.r.t. perturbations in the inputs, and in
%fact the effect of  an instantaneous perturbations in the input decays exponentially fast.
%The requirement that the output function is Lipschitz ensures that the
%output trajectories inherit the favorable robustness properties of the state trajectories.
%either a linear or  %output function or apply 
%one of the standard activation functions %, which are Lipschitz.
%
%\begin{Remark}[Role of constants in robustness]
%\label{rem:constants}
%    The smaller the constants $C$, $\tau$, $L$  and $L_{g,1},L_{g,2}$ are, the more robust the system is to perturbations in inputs and states. 
%    Among these constants, $\tau$  is the most significant one, as  it determines how fast the effect of perturbations will decay with time. % and it is the most significant measure of robustness. 
%\end{Remark}
UEC implies existence of a unique stable stationary solution on $\mathbb{Z}$, UEIIS implies continuity in initial states and inputs, and Lipschitzness of the readout map extends these properties to outputs. By \cite[Theorem 12]{TanTAC}, UEC implies UEIIS under local Lipschitz continuity and uniform convergence on compact invariant sets.
%\begin{color}{black}
Stationary solutions are also continuous in input trajectories:
\begin{lemma}
\label{lemma:oldNewS}
    If $\mathcal{S}$ is class \class{}, then
    for any two bounded input trajectories $\vv_1,\vv_2:\sZ\to \sV$,\begin{align*}
    %\label{def:classDef:mem}
        \|\vs_{\vv_1}(t)-\vs_{\vv_2}(t)\|_2\leq L\sum_{k=1}^\infty \tau^{k-1}\|\vv_1(t-k)-\vv_2(t-k)\|_2
    \end{align*}
\end{lemma}
In particular, Lemma \ref{lemma:oldNewS} implies 
that  every system of class \class{} is also of class \classOld{} as defined in \cite{eringis2023pacbayesRNN}, see \cite[Corollary 3.1]{arxiv}.
In addition, outputs of \class{} systems generated by mixing and stationary inputs are also mixing and stationary, 
see \cite[Lemma A4.1]{eringis2023pacbayesRNN_OLD}.
A sufficient condition for a system to be of class \class{}  \textcolor{black}{which will be used in this paper} is that $f$ and $g$ are Lipschitz, and that there exists a quadratic Lyapunov function 
\begin{lemma}
\label{lemma:P:classS}
   Assume that $f:\mathbb{R}^{n} \times \mathbb{R}^{m} \to \mathbb{R}^{n}$  is continuously differentiable in the first argument, and $L_f$-Lipschitz continuous and $g:\mathbb{R}^{n} \times \mathbb{R}^{m} \to \mathbb{R}^{p}$ satisfies Definition \ref{def:classDef}, and
   $f(0,0)=0$, $g(0,0)=0$.
   Assume that $C_u >0$, $0 <\tau < 1$ are such that there exists a matrix $P > 0$ such that for every $s\in\mathbb R^n$ and for all $\vv \in \mathbb{R}^{m}$, $\|\vv\|_{2} \le C_u$
     \begin{equation}
      \label{P:classS}
J_s f(s,v)^\top P J_s f(s,v)\preceq \tau^2P.
  %   (\dfrac{f(\vs,\vv)}{d\vs})^TP(\dfrac{f(\vs,\vv)}{d\vs})- \tau^2 P.
     \end{equation}
   Then $\mathcal{S}=(\mathbb{S},\mathbb{V},\mathbb{Y},f|_{\mathbb{S}\times \mathbb{V}},g|_{\mathbb{S}\times \mathbb{V}})$ is a class \class{} system with constant $C,\tau,L,L_{g,1},L_{g,2}$, such that
   Then define,
   \[
   \begin{split}
    & L=L_{f}\frac{\sqrt{\lambda_{max}(P)}}{\sqrt{\lambda_{min}(P)}}, ~ 
    C=\sqrt{\frac{\sqrt{\lambda}_{max}(P)}{\lambda_{min}(P)}} \\
    & \mathbb{S}=\{x \in \mathbb{R}^{n} \mid \sqrt{x^TPx} \le \frac{\sqrt{\lambda_{min}(P)}LC_u}{1-\tau}\} \\
   & \mathbb{V}=\{v \in \mathbb{R}^{m} \mid \|v\|_{2} \le C_u\} \\
   & \mathbb{Y}=\{y \in \mathbb{R}^{p} \mid \|y\|_{2} \le 
    \left(\frac{L_{g,1}L}{(1-\tau)}+L_{g,2}\right)C_u\} 
   \end{split}
  \]
\end{lemma}
The proof of Lemma \ref{lemma:P:classS} is based on \cite[Theorems 8, 13]{TanTAC,TranCDC}.
In particular, Lemma \ref{lemma:P:classS} shows that stable LTI systems restricted to bounded invariant sets form class \class{} systems.
For piecewise-affine functions $f$ (e.g., RNNs with ReLU activation) or sector-bounded nonlinearities,
the conditions in \eqref{lemma:P:classS} can be verified using Linear Matrix Inequalities \cite{PavlovTac2011,TanTAC}.
Since cascades of class \class{} systems are again class \class{} \cite[Lemma \mref{lemma:series}]{eringis2023pacbayesRNN_OLD},
many relevant systems satisfy this property.

% %\par
% For instance, %A large class of such examples are %\ref{lemma:classExamples}, 
% stable linear state-space systems, i.e., systems of the form
% $f(\rvs,\rvw)=A\rvs+B\rvw$ and
% $g(\rvs,\rvw)=C\rvs+D\rvw$
% for suitable matrices $A,B,C,D$ such that the spectral radius of $A$ is smaller than $1$, satisfy 
% 	\cite[Lemma \mref{lemma:classExamples}]{eringis2023pacbayes}, but they
% are not contractions. 
%such systems are of class \class{}, if
%$A$ is Schur, i.e., its spectral
%radius $\rho(A)$  is smaller than $1$. It is well-known in control theory \cite{Katayama:05} that then condition 3 of Lemma \ref{lemma:classExamples} holds. 
%However, $f$ is not necessarily a contraction. 
%\par
% For piecewise-affine functions $f$
% (e.g., RNNs with ReLu activation)
% 	the conditions of \cite[Lemma \mref{lemma:classExamples}]{eringis2023pacbayesRNN_OLD}
% can be checked 
% %which is the case for RNNs with a ReLu activation, 
% %%such Lyapunov functions can be found 
% using Linear Matrix Inequalities (LMI), see \cite[Theorem 2]{PavlocTac2011}.

%Class \class{} property is preserved under interconnection:
%series interconnection of two class \class{} systems is also \class{} 
%(see \cite[Lemma \mref{lemma:series}]{eringis2023pacbayesRNN_OLD}). Furthermore, the corresponding constants can be computed from those of 
%the two systems. 
%In particular, multilayer RNNs which can be represented as a  series interconnection of several single layer RNNs will be of class \class{}, if each individual layer is of class \class{}.

\section{Assumptions on Data and Predictors}\label{sec:ProblemForm} 
%Let us now formally state our assumptions on the data and the class of predictors.
%the learning problem. %Firstly we assume that the processes involved are essentially bounded, i.e. $\|[\rvy^T(t),\rvw^T(t)]\|_\infty<\infty$
%Formally, we assume that the processes $\rvy,\rvw$ can be realised by a class \class{} system driven by i.i.d. noise, and the training data $\train=\{\rvy(s),\rvw_\rve(s)\}_{s=0}^{N-1}$ are samples from the unique stationary solution of a class \class{} system driven by i.i.d. noise. 
%Recall that $\rvy,\rvw$ two stochastic processes, where 
%$\rvy(t)$ is the process we wish to estimate from present and past, $\{\rvw(s)\}_{s=0}^t$ of process $\rvw$. 
\par
%\textcolor{black}{Conc}
Informally, we assume that $\rvy$ and $\rvu$ are outputs of a class \class{} system for an i.i.d. input process, and the predictors to be class \class{}
systems.  To this end, we need to define the output process of  class \class{} systems driven by stochastic inputs. Let $S=(\sS,\sY,\sV,f,g)$ be a system of class \class{}, and let $\rvv$ be an essentially bounded stochastic process taking values in $\sV$.
%Then for any $\omega \in \Omega$,
%the sample path $\rvv(\omega): \sZ \ni s \mapsto \rvv(s)(\omega) \in \sV$ is a bounded input trajectory. 
%\par
For any initial state 
$\vs_0$, time instances $t_0$, we define the 
random variables 
$\rvy_S(\vs_0,t_0,\rvv;t):\omega\mapsto
\rvy_S(\vs_0,t_0,\vv;t)$, $\vv(s)=\rvv(s)(\omega)$, and
%and
%$\Omega \ni \omega \mapsto \rvs_S(\vs_0,t_0,\rvv(\omega);t)$
%is random variables, which we will denote by $\rvy_S(\vs_0,t_0,\rvv;t)$ and 
we refer to them 
%$t \mapsto \rvy_S(\vs_0,t_0,\rvv;t)$  
as the  \emph{output process of $S$  at time $t$ 
for the stochastic input $\rvv$, initial state $\vs_0$ and initial time $t_0$.}
%call the \emph{output  of $S$
%$\rvs_S(\vs_0,t_0,\rvv;t)$ and refer to as the
%\emph{output and state of $S$  at time $t$ for the stochastic input $\rvv$, initial state $\vs_0$ and initial time $t_0$.}
%Note that for all $\sZ \ni t \ge t_0$
%\begin{align*}
%&\rvs_S(\vs_0,t_0,\rvv;t+1)=f(\rvs_S(\vs_0,t_0,\rvv;t),\rvv(t)), ~ \quad \rvs_S(\vs_0,t_0,\rvv,t_0)=\vs_0, \\
% &   \rvy_S(\vs,t_0,\rvv;t)=g(\rvs_S(\rs_0,t_0,\rvv,t),\rvv(t))
%\end{align*}
%\par
Likewise, 
%there exist
%unique steady-state output and state trajectories $\rvy_{S,\rvv(\omega)}$ and
%$\rvs_{S,\rvv(\omega)}$
%of $S$ associated with the input %$\rvv(\omega)$.
for every $t \in \sZ$, we define the random variables $\rvy_{S,\rvv}(t):\omega\mapsto\rvy_{S,\vv}(t)$, $\vv(s)=\rvv(s)(\omega)$ 
%and  $\Omega \ni \omega \mapsto \rvs_{S,\rvv(\omega)}(t)$
%defines a random variable
%which we denote by 
%M$\rvy_{S,\rvv}(t)$ 
and call them
%$\rvs_{S,\rvv}(t)$ respectively, and call them
%In particular, for $\rvs(t)=\rvs_S(\rs_0,t_0,\rvv,t)$ respectively $\rvs(t)=\rvs_{S,\rvv}(t)$, it holds
%that $\rvs(t+1)=f(\rvs(t),\rvv(t))$ for all $t \ge t_0$ respectively all $t \in sZ$.
%Moreover, $rvy_S(\vs,t_0,\rvv;t)=g(\rvs_S(\rs_0,t_0,\rvv,t),\rvv(t)$ and
%$rvy_{S,\rvv}(t)=g(\rvs_{S,\rvv}(\rvs(t),\rvv(t))$. 
%We call the stochastic processes $\rvy_{S,\rvv}$ and $\rvs_{S,\rvv}$
\emph{the steady-state output process 
of $S$ associated with $\rvv$.}
We are now ready to state our assumptions
on the data. 
%In particular, we can define the random
%\textcolor{black}{We assume the following}
\begin{Assumption}[Data Generator]
\label{as:generator}
The system \eqref{eq:gen}, i.e., the system  $S_g=(\sS_g,\sW, \sV_g,f_g,g_g)$, is of class \class{}, with constants 
$C_g,\tau_g,L_g,L_{g_g,1},L_{g_g,2}$, and $\begin{bmatrix} \rvy^T, & \rvu^T\end{bmatrix}^T$ is the steady-state output process of $S_g$, i.e.,
%$C_g=C_g(S_g)$, $\tau_g=\tau_g(S_g)$, $L_{g}=L_{g}(S_g)$, $L_{g_g,1}=L_{g_g,1}(S_g)$, $L_{g_g,2}=L_{g_g,2}(S_g)$ 
\[ \begin{bmatrix} \rvy^T(t), & \rvu^T(t)\end{bmatrix}^T=\rvy_{S_g,\rve_g}(t).  \] 
%and an essentially bounded i.i.d. process $\rve_g$
%such that 
%$\begin{bmatrix} \rvy^T,\rvw^T \end{bmatrix}^T$ is the steady-state output process
%of $S_g$ associated with
%$\rve_g$, i.e., $\begin{bmatrix} \rvy^T,\rvw^T \end{bmatrix}^T=\rvy_{S_g,\rve_g}$.
%\[ 
%\begin{bmatrix} \rvy(t)(\omega) \\ \rvw(t)(\omega) \end{bmatrix}=\rvy_{S_g,\rve_g(\omega)}(t), \quad \quad t \in \sZ, \omega  \in \Omega
%\]
%where $\rve_{g}(\omega):\sZ \ni s \mapsto \rve_g(s)(\omega) \in \sV$ is the sample path of the process $\rve_g$ for the event $\omega \in \Omega$.
%The data $(\rvy(t),\rvw(t))$, can be realised as an output of a class \class{} system, with $\rvv=\rve_g$ i.i.d., i.e.,
%\begin{subequations}\label{eq:generatorSys}
%    \begin{align}
%    \rvs_g(t+1)=f_g(\rvs_g(t),\rve_g(t))\\
%    \begin{bmatrix} \rvy(t)\\\rvw(t)\end{bmatrix} = h_g(\rvs_g(t),\rve_g(t))
%\end{align}
%\end{subequations}
%Furthermore we assume, $(f_g,h_g)$ of class \class{}, $\rve_g$ is i.i.d. and essentially bounded, and finally $\rvs_g=\rvs_{\rve_g}$, i.e., $\rvs_g$ is the bounded (stationary) solution, as defined in Definition \ref{def:classDef}, exponential convergence property.
\end{Assumption}
%Note that, we assume the existence of the generator $S_g$, but we do not assume the knowledge of $S_g$ when performing the system identification.
%Assumption \ref{as:generator} is a  realisability assumption, as the data generator is of the same type as the predictor.
Note that, we assume the existence of the generator $S_g$, but we do not assume the knowledge of $S_g$ when performing the system identification.
%Assumption \ref{as:generator} is a  realisability assumption, as the data generator is of the same type as the predictor.
%uses dynamical systems of class \class{} for prediction, it is natural to assume that the data is generated by such systems too. 
Assumption \ref{as:generator} guarantees certain mixing  properties of $\rvw,\rvy$ \cite[Lemma A4.1]{eringis2023pacbayesRNN_OLD} which, in turn, allows the use of concentration inequalities \cite{rio2000inegalites}.
We assume that the systems \eqref{eq:dyn1} are class \class{}:
%and they are parametrised by elements of a parameter set $\Theta$. 
%\begin{color}{black}
	\begin{Assumption}[Parameterisation] \label{as:parametrisation}
%Assume that 
 The system $\mathcal{S}_{p}=(\sS,\sW \times \Theta, \sY,\hat{f},\hat{g})$ is a system of class \class{} with constants 
 \[ (\bar{C},\tau,\bar{L},\bar{L}_{g,1},\bar{L}_{g,2})\!=\!C(\mathcal{S}_{p},\tau(\mathcal{S}_{p}),L(\mathcal{S}_{p}),
 L_{g,i}(\mathcal{S}_{p}), L_{g,i}(\mathcal{S}_{p})) \]
 and \textbf{(1)} $\hat{f},\hat{g},\hat{\vs}_s$ are continuous, \textbf{(2)} $\hat{\vs}_s$ is $L_s$ Lipschitz, \textbf{(3)} 
 %$\Pi(\theta)(\{\rvw(s)\}_{s=0}^{t})
 $\hat{\rvy}_{\theta}(t\! \mid\! 0)=\rvy_{\mathcal{S}_{p}}(\hat{\vs}_s(\theta),0,(\rvw,\theta);t)$, \textbf{(4)} $\hat f(0,0,\theta)=0$ and $\hat g(0,0,\theta)=0$ for all $\theta\in\Theta$.
 \end{Assumption}
 \begin{Assumption} \label{as:loss} The loss function $\ell$
is Lipschitz, with global Lipschitz constant $L_\ell>0$ %i.e.,
\end{Assumption}
Note that Assumption \ref{as:parametrisation}
implies that for any \textcolor{black}{parameter}
$\theta \in \Theta$, 
the output $\hat{\rvy}_{\theta}(t|0)$ %of the hypothesis
\textcolor{black}{is} bounded as $\mathbb{Y}$ is bounded.
% i.e., $\|\hat{\rvy}_{\theta}(t|0)\| \le G_\theta B^{\rvq}$,
%where $G_\theta\triangleq \frac{L_{g,1}(\theta)L_g(\theta)}{1-\tau(\theta)}+L_{g,2}(\theta)$.
%Note that, since we consider bounded processes, one can 
Since all the processes are bounded, one can use \textcolor{black}{restrictions} of locally Lipschitz loss functions, e.g. square loss, 
\textcolor{black}{to a bounded set, and they will be Lipschitz.}
 \begin{lemma}\label{lemma:NewSI} 
   Under Assumption \ref{as:parametrisation}-\ref{as:loss} the following holds.
   
  \textbf{Class \class{} predictors.} For every $\theta \in \Theta$, if we set 
   \begin{align*}
	&	\hat{f}_{\theta}:  \sS \times \sW \ni (\vs,\vv) \mapsto \hat{f}(\vs,\vv,\theta) \in \sS, \\
	&	 \hat{g}_{\theta}: \sS \times \sW \ni (\vs,\vv) \mapsto \hat{g}(\vs,\vv,\theta) \in \sY
\end{align*}
then $S_{\theta} \triangleq (\sS,\sY,\sW,\hat{f}_{\theta},\hat{g}_{\theta})$ is a system of class \class{}
\begin{color}{black} with %constants
\end{color}
\begin{align*}
    & (C(S_{\theta}),\!\tau(S_{\theta}),\!L(S_{\theta}),\!L_{g,1}(S_{\theta}),\!L_{g,2}(S_{\theta}))\!=\!(\bar{C},\!\bar{\tau},\!\bar{L},\!\bar{L}_{g,1},\!\bar{L}_{g,2}),
\end{align*}
and $\hat{\rvy}_{\theta}(t\!\mid\! t_0)$ is the output of $S_{\theta}$ from initial state $\hat{\rvs}_{s}(\theta)$ and initial  time $t_0$, and  $\hat{\rvy}_{\theta}(t)$ is the stationary output of $S_{\theta}$
for the input $\rvw=\begin{bmatrix} \rvy^T, & \rvu^{T} \end{bmatrix}^T$, i.e., 
\begin{align*}
    & \hat{\rvy}_{\theta}(t\! \mid\! t_0)=\rvy_{S_{\theta}}(\hat{\rvs}_{s}(\theta),t_0,\rvw;t), \quad  \hat{\rvy}_{\theta}(t)=\rvy_{S_{\theta},\rvw}(t).
\end{align*}
\begin{color}{black}
 \textbf{Lipschitzness of the predictors.}\end{color}
 %\textcolor{red}{TODO check L_s}
  The maps $\theta \mapsto \rvy_\theta(t \mid 0)$ and $\theta \mapsto \rvy_\theta(t)$ have a Lipschitz constant 
   \[L_{\text{pred}}=\bar{L}_{g,1} \max\{L_{\rs} \bar{C}, \frac{\bar{L}}{1-\bar{\tau}}\} + \bar{L}_{g,2}, \]
  %$L_{\text{pred}}$, i.e.,
  i.e., %$L_{\text{pred}}$ such that 
  for all $\theta_1,\theta_2 \in \Theta$, 
  \begin{equation} 
  \label{as:pred_lipschitz}
     \begin{aligned}
      \|\rvy_{\theta_1}(t\mid 0) - \rvy_{\theta_2}(t\mid 0)\| \le L_{\text{pred}} \|\theta_1 - \theta_2\| \\
      \|\rvy_{\theta_1}(t) - \rvy_{\theta_2}(t)\| \le L_{\text{pred}} \|\theta_1 - \theta_2\| \\
  \end{aligned}
\end{equation}
\textbf{Existence of true loss} \eqref{limits_true_loss} holds and hence the true loss \eqref{gen:losses} is well-defined.
 \end{lemma}
 The first two statements follow from the properties of class \class{} systems, the last statement is based on
 \cite[Lemma \mref{lemma:stationary_x_v}]{eringis2023pacbayesRNN_OLD}. %, the proof is in the appendix. %\cite{Arxiv}.
Finally, we need the some geometric regularity assumptions on the parameter set $\Theta$. To this end,
let $B_{\Theta}$ be the $\sigma$-algebra of Lebesgue-measurable subsets of the parameter set $\Theta$, and $m$ denote the Lebesgue measure on $\mathbb{R}^{n_\theta}$, and for any Lebesgue-measurable subset  $B \subseteq \mathbb{R}^{n_\theta}$,
let $\operatorname{vol}(B) = m(B)$. For any $\theta \in \mathbb{R}^{n_\theta}$, denote by $B(\theta,r)$ the ball of radius $r$ centered at $\theta$, i.e., $B(\theta,r) = \{\theta' \in \Theta : \|\theta' - \theta\| \le r\}$.
\begin{Assumption}[Geometric regularity of the parameter set]
  \label{as:parameter_set}
  Assume that $\Theta$ is compact, $\operatorname{vol}(\Theta)>0$, and 
  %let $\pi$ be the uniform probability measure on $\Theta$, i.e., 
  %$\pi(B) = \frac{\operatorname{vol}(B)}{\operatorname{vol}(\Theta)}$ for any measurable set $B \subseteq \Theta$.
  that there exists $c>0$ such that $\operatorname{vol}(B(\theta,r) \cap \Theta)\ge c\,r^{n_\theta}\mathrm{vol}(\Theta)$ 
  for every $\theta\in\Theta$ and $0<r\le1$.
\end{Assumption}
%This assumption avoids some geometric irregularities on the boundary of $\Theta$.
From \cite[Proof of Lemma E22]{Nips2026} it follows that if $\Theta$ is a ball of radius $R > 2$ centered at $\theta_m$, i.e., $\Theta = B_{R}(\theta_m)$, then Assumption \ref{as:parameter_set} is satisfied with 
$c=\frac{\sqrt{\pi}^{n_{\theta}} \Gamma(n_{\theta}/2+1)}{2^{n_{\theta}}\operatorname{vol}(\Theta)}$, where $\Gamma$ denotes the $\Gamma$ function.

\section{Main result}\label{sec:Results}
The main result is as follows. 
%With the definitions and assumptions above, we can state the main result.
% To this end,  
% %for any probability density $\pi$ on $(\Theta,B_{\theta},m)$, denote by $\mathcal{M}_\pi$ the set of all probability densities for which the corresponding probability measures are absolutely continuous w.r.t. to the probability measure defined by $\pi$. 
% %Furthermore, 
% for any $\theta \in \Theta$, denote by $C(\theta),\tau(\theta),L_{g}(\theta),L_{g,1}(\theta),L_{g,2}(\theta)$
% 	the constants of the class \class{}  system $\mathcal{S}_{\theta}$, and 
% 	 $\hat{\vs}(\theta)$ be the initial state of $S_{\theta}$ from which the predictor $h_{\theta}$ is generated, see Assumption
% 	 \ref{as:parameterisation}. 
% 	Let $\bar{\theta}_{\infty}^\rvq(1)$ and $B^\rvq$ be  the constants of	the data generating system from Lemma \ref{as:gen_mixing:lem}.

% \textcolor{black}{
\begin{Theorem}[PAC bound]\label{thm:mainThm} %Let predictors $h\in\mathcal{H}$\deividas{, see \eqref{eq:dyn1}}, be realised by \class{} class dynamical systems (see Def. \ref{def:classDef}), and be parameterised by weights $\theta\in\Theta\subset \sR^{n_\theta}$ (see Assumption \ref{as:parameterisation}),

%	with associated system constants $C(\theta),L_{g,1}(\theta),L_{g}(\theta),L_{\rvv}(\theta)$\deividas{, and $\tau(\theta)\in[0,1)$}. 
	Under Assumptions \ref{as:generator},\ref{as:parametrisation}, \ref{as:loss}, \ref{as:pred_lipschitz}, and \ref{as:parameter_set},
  % Let $G_1,G_2$ be the uniform constants from \cite{eringis2023pacbayesRNN_OLD}[def. (10) and eq. (117)], applied to the predictor family $\{\mathcal S_\theta:\theta\in\Theta\}$, with $L_v(\theta)=L(\mathcal S_\theta)$, $L_{g,s}(\theta)=L_{g,1}(\mathcal S_\theta)$, and $L_{g,v}(\theta)=\sqrt{1+L_{g,2}^2(\mathcal S_\theta)}$.
	for any $\delta\in(0,0.5]$, and $N \ge 3$, 
    \begin{multline} \label{eq:mainThmIneq}
         \bP\left(\forall  h \in \mathcal{H}: |\mathcal{L} (h)-\hat{\mathcal{L}}_N(h)| \le \epsilon_N(\delta)
         \right) \ge 1-2\delta
    \end{multline}
    where the constants are defined as follows:
\begin{equation}\label{eq:G1G2}
\begin{split}
    \epsilon_N(\delta) &\triangleq 2\frac{(\ln\dfrac{2}{\delta}	+ C_{\text{pac}}) \ln(N)}{\sqrt{N}} \\
    C_{\text{pac}} &\triangleq 
    L_{\ell}L_{\text{pred}} + \log(1/c) % \ln\left(\frac{vol(\Theta)}{\sqrt{\pi \sigma^2}^{n_{\theta}} \Gamma(n_{\theta}/2+1)}\right)\\
    %&\qquad 
    + \frac{n_{\theta}}{2} + 
     \frac{(G_1+G_2)}{2} \\
    G_1 & \triangleq L_{\ell}\left( B^{\rvq}\left(\bar{G}+\bar{H} \right) +\bar{\theta}_{\infty,N}^{\rvq}(1) \bar{G} \right)^2 \\
    G_2 & \triangleq 2L_{\ell}\bar{C} \left( 2 B^\rvq \bar{H} +\sup_{\theta\in\Theta}\|\hat{\vs}_{s}(\theta)\|\frac{\bar{L}_{g,1}}{1-\bar{\tau}} \right) \\
    %\triangleq \sup_{\theta\in\Theta}2 L_\ell^2\left (B^\rvq \left ( G_\theta+ H_\theta \right )+\bar{\theta}_{\infty,N}^{\rvq}(1) G_\theta \right )^2,\\
	%G_2 &\triangleq \sup_{\theta\in\Theta}2 L_\ell C(\theta)  \\œ
  \bar{G} &\triangleq \frac{\bar{L}_{g,2}\bar{L}_{g,1}}{1-\bar{\tau}}+\bar{L}_{g,2},\; \bar{H}\triangleq \frac{\bar{L}_{g,1} \bar{L}_{g,2}}{(1-\bar{\tau})^2} \\
  B^{\rvq} & \triangleq 2C_{e_g} \left ( L_{g_g,2}+\frac{L_{g}L_{g_g,1}}{1-\tau_g}\right ),  \\
  \bar{\theta}_{\infty}^{\rvq}(1) & \triangleq 2C_{e_g} \frac{L_{g}L_{g_g,1}}{(1-\tau_g)^2}, \quad 
   C_{e_g} \triangleq \operatorname*{ess\,sup} \|e_g(t)\|_2
\end{split}
\end{equation}
%where 
%µ$ and with $vol(\Theta)$ the volume of $\Theta$, $\Gamma(\cdot)$ the gamma function. % and $H_\theta$ the transfer function corresponding to $S_\theta$.
\end{Theorem}
Most constants depend on the predictor class, except $(C_g,\tau_g,L_g,L_{g_g,1},L_{g_g,2})$, which depend on the data generator and can be bounded from partial knowledge of it, see \cite{eringis2023pacbayesRNN}. Intuitively, $B^\rvq$ and $G_2$ bound the norm of data and predictions, while $\bar{\theta}_{\infty}^{\rvq}(1)$ and $G_1$ bound their mixing degree. For i.i.d. data, $\bar{\theta}_{\infty}^{\rvq}(1)=0$. The bounds decrease with the Lyapunov exponents $\tau_g$ and $\bar{\tau}$ of the data generator and predictor: greater stability yields tighter bounds. 
%\cite{eringis2023pacbayesRNN} for more details. 
\begin{pf}[Sketch of the proof]
 For any probability density $\pi$ on $(\Theta,B_{\theta},m)$, denote by $\mathcal{M}_\pi$ the set of all probability densities for which the corresponding probability measures are absolutely continuous w.r.t. the probability measure defined by $\pi$. Moreover, if $g:\mathcal{H} \rightarrow \mathbb{R}$ is a map such that $\theta \mapsto g(h_{\theta})$ is measurable and absolutely integrable, then %we define
\begin{equation*}
	E_{\rho} g\triangleq\int_{\theta \in \Theta} \rho(\theta)g(h_\theta)dm(\theta)  %\label{eq:Edef}
\end{equation*}
denotes the expectation of $\theta \mapsto g(h_{\theta})$ w.r.t. the density $\rho$. Hence the expectations $E_{\rho} \hat{\mathcal{L}}_N$ and  $E_{\rho} \mathcal{L}$ are well defined.

  From \cite{eringis2023pacbayesRNN}\!\! we get the %following 
  PAC-Bayesian inequality
  \begin{multline} \label{eq:pac:gen1}
        \bP(\forall\hat{\rho}\in\mathcal{M}_\pi: E_{\hat{\rho}} \mathcal{L} - E_{\hat{\rho}} \hat{\mathcal{L}}_N \le \\
           \dfrac{1}{\lambda}\!\left[ \KL(\hat{\rho} \|\pi) +
	\ln\dfrac{1}{\delta}+ %\frac{1}{\sqrt{N}}  
  \frac{\lambda^2}{N}  G_1 +\frac{\lambda}{N}G_2  \right]) \ge 1-2\delta
        %r_N(\lambda,\delta,\hat{\rho},\pi) 
    \end{multline}

  First, we take $\lambda=\sqrt N$ and $\pi=\frac{1}{vol(\Theta)}$ to be the uniform distribution.
For $r=N^{-1/2}$, let $\rho_\theta$ be uniform on $B(\theta,r)\cap\Theta$. Then
    \[
    D_{\rm KL}(\rho_\theta\Vert\pi)
    =\log\frac{1}{\pi(B(\theta,r))}
    \le \log(1/c)+\frac{n_\theta}{2}\log N.
    \]
  %For any $\theta \in \Theta$ we take $\rho_\theta$ to be the prior centered around %$\theta$: 
  %$$\rho_{\theta}(\theta')=\left\{\begin{array}{rl}
  %     \frac{1}{vol(B(\frac{1}{\sqrt{N}}))} & \text{if } \theta' \in B(\frac{1}%{\sqrt{N}}) \\
      % 0 & \text{otherwise}
     %\end{array}\right.$$    
     %where $B(\frac{1}{\sqrt{N}})$ is the ball of radius $\frac{1}{\sqrt{N}}$ centered %at $\theta$.
%and %Similarly to \cite[Lemma C.24]{Nips2026} 
%\begin{multline}
%\KL(\rho \|\pi_\theta) \le 
%  \ln\left(\frac{vol(\Theta)}{vol(B(\frac{1}{\sqrt{N}}))}\right)= \\
%  \left(\ln\left(\frac{vol(\Theta)}{\sqrt{\pi \sigma^2}^{n_{\theta}} %\Gamma(n_{\theta}/2+1)}\right)+ \frac{n_{\theta}}{2} \right) \ln(N)
%\end{multline}
From \ref{as:pred_lipschitz} it follows that
$\theta \mapsto \mathcal{L}(\theta)$ and 
$\theta \mapsto \hat{\mathcal{L}}_N(\theta)$ are $L_{\ell}L_{\text{pred}}$-Lipschitz. By repeating the argument of the proof
of \cite[Corollary  C23]{Nips2026}%, it follows that
\begin{equation}
  \label{loss:lipschitz1}
\begin{split}
   & \mathcal{L}(\theta) \le E_{\rho_\theta} \mathcal{L} \!+\!
     L_{\ell}L_{\text{pred}} E_{\rho_\theta}\psi_\theta \le  
      E_{\rho_\theta} \mathcal{L} \!+\! \frac{L _{\ell}L_{\text{pred}}}{\sqrt{N}} \\
    & \mathbb E_{\rho_\theta}\hat L_N
\! \le\! \hat L_N(\theta)\!+\!L_\ell L_{\rm pred}\mathbb E_{\rho_\theta}\psi_\theta
\le \hat L_N(\theta)\!+\!\frac{L_\ell L_{\rm pred}}{\sqrt N}
\end{split}
\end{equation}
with $\psi_\theta(\theta') \triangleq\|\theta' - \theta\|_2 $ and that
\begin{multline}
  \label{eq:pac_un1}
  \bP(\forall \theta \in \Theta: \mathcal{L}(\theta) - \hat{\mathcal{L}}_N(\theta) \le
  \\ \frac{(\ln(1/\delta)+ 2C_{\text{pac}}) \ln(N)}{\sqrt{N}}) \ge 1-2\delta
\end{multline}
Similarly, by repeating the proof of \eqref{eq:pac:gen1} with obvious modifications, we obtain
\begin{multline} \label{eq:pac:gen2}
        \bP\Big(\forall\hat{\rho}\in\mathcal{M}_\pi: E_{\hat{\rho}} \hat{\mathcal{L}} - E_{ \hat{\rho}}\mathcal{L} \le \\
        \dfrac{1}{\lambda}\!\left[ \KL(\hat{\rho} \|\pi) +
	\ln\dfrac{1}{\delta}+ %\frac{1}{\sqrt{N}}  
  \frac{\lambda^2}{N}  G_1 +\frac{\lambda}{N}G_2  \right]\Big) \ge 1-2\delta
        %r_N(\lambda,\delta,\hat{\rho},\pi) 
  \end{multline}
    By using 
    the counterpart of
    \eqref{loss:lipschitz1}, we get for any $\theta \in \Theta$
\begin{equation}
\begin{split}
    \label{loss:lipschitz2}
       &\hat{\mathcal{L}}_N(\theta) \le E_{\rho_{\theta}} \hat{\mathcal{L}}_N +L_{\ell}L_{\text{pred}} E_{\rho_{\theta}}\psi_\theta\\
       &\qquad\qquad\le 
       E_{\rho_{\theta}} \hat{\mathcal{L}}_N + L_{\ell}L_{\text{pred}}\frac{1}{\sqrt{N}} \\
      &E_{\rho_{\theta}} \mathcal{L} \le \mathcal{L}(\theta) + L_{\ell}L_{\text{pred}}\frac{1}{\sqrt{N}}
    \end{split}
\end{equation}
    and by repeating the arguments in the proof of \eqref{eq:pac_un1}, it follows that
    \begin{multline}
      \label{eq:pac_un2}
      \bP\left(\forall \theta \in \Theta: \hat{\mathcal{L}}_N(\theta) - \mathcal{L}(\theta) \le \frac{(\ln(1/\delta)+ 2C_{\text{pac}}) \ln(N)}{\sqrt{N}}\right)\\ \ge 1-2\delta
    \end{multline}
    Taking $\delta/2$ instead of $\delta$ in \eqref{eq:pac_un1} and \eqref{eq:pac_un2}
    and using the union bound, the statement of the theorem follows. 
\end{pf}
Next, we derive an inequality to compare the true loss of the PEM model.
% $\varepsilon_N(\delta)
%:=2\left(\log\frac{2}{\delta}+C_{\rm pac}\right)
%\frac{\log N}{\sqrt N}$ 
\begin{Theorem}[Oracle inequality]
\label{thm:oracle_inequality}
 For $\hat\theta_N\in\arg\min_{\theta\in\Theta}\hat{\mathcal{L}}_N(\theta)$, with the assumptions of Theorem \ref{thm:mainThm},
 for all $\delta \in (0,0.5)$:
  \[
    \bP\left(\mathcal{L}(\hat \theta_N) \le \inf_{\theta \in \Theta} \mathcal{L}(\theta) +2\varepsilon_N(\delta)\right) \ge 1-2\delta
  \]
  %\[
  %  \bP\left(\mathcal{L}(\theta_N) \le \inf_{\theta \in \Theta} \mathcal{L}(\theta) + \frac{(\ln\dfrac{2}{\delta}	+ 2C_{\text{pac}}) \ln(N)}{\sqrt{N}}\right) \ge 1-\delta
  %\]
\end{Theorem}
\begin{pf}[Proof of Theorem \ref{thm:oracle_inequality}]
By Theorem~\ref{thm:mainThm} we have $\| \mathcal{L}(\theta)-\hat{\mathcal{L}}_N(\theta)\| \le \varepsilon_N(\delta)$ w.r.t. $1-2\delta$ for all $\theta\in\Theta$ thus 
$\mathcal{L}(\hat{\theta}_N) \le \hat{\mathcal{L}}_N(\hat{\theta}_N)+\varepsilon_N(\delta)$ and by the definition $\hat\theta_N\in\arg\min_{\theta\in\Theta}\hat{\mathcal{L}}_N(\theta)$ so $\hat{\mathcal{L}}_N (\hat \theta_N )\leq \hat{\mathcal{L}}_N (\theta)$ therefore $\mathcal{L}(\hat\theta_N) \le \hat{\mathcal{L}}_N(\theta)+\varepsilon_N(\delta)\le \mathcal{L}(\theta) +2\varepsilon_N(\delta)$ where we used that $\hat{\mathcal{L}}_N(\theta) \le \mathcal{L}(\theta)+\varepsilon_N(\delta)$ by ineq.~\eqref{eq:pac_un2}. Finally, we take the infimum over $\theta\in\Theta$ to prove the theorem.
  %  Consider the event 
  %  \begin{multline*}
  %   A=\Big\{ \omega \in \Omega \mid \forall \theta: |\mathcal{L}(\theta) - \hat{\mathcal{L}}_N(\theta)(\omega)| \\ \le \frac{(\ln\dfrac{2}{\delta}+ 2C_{\text{pac}}) \ln(N)}{\sqrt{N}} \Big\}
  %  \end{multline*}
  %  Note that for any $\omega \in A$, %for any $h \in \mathcal{H}
  %  \begin{align}
  %     & \mathcal{L}(\theta_N(\omega)) \le \hat{\mathcal{L}}_N(\theta_N(\omega))(\omega) + \frac{(\ln\dfrac{2}{\delta}+ 2C_{\text{pac}}) \ln(N)}{\sqrt{N}} \label{pf:oracle1} \\
  %     & \forall \theta \in \Theta: \hat{\mathcal{L}}_N(\theta)(\omega) \le 
  %     \mathcal{L}(\theta) + \frac{(\ln\dfrac{2}{\delta}+ 2C_{\text{pac}}) \ln(N)}{\sqrt{N}} \label{pf:oracle2}
  %  \end{align}
  %  Since $\hat{\mathcal{L}}_N(\theta_N(\omega))(\omega) \le \hat{\mathcal{L}}_N(\theta)(\omega)$ for all $\theta \in \Theta$, combining \eqref{pf:oracle1} and \eqref{pf:oracle2} yields 
  %  \begin{equation}
  %   \label{pf:oracle3}
  %     \forall \theta \in \Theta: \mathcal{L}(\theta_N(\omega)) \le \mathcal{L}(\theta) + 2\frac{(\ln\dfrac{2}{\delta}+ 2C_{\text{pac}}) \ln(N)}{\sqrt{N}}
  %  \end{equation}
  %  That is, the event $A$ is contained in the event 
  %  \[
  %  \left\{\omega \in \Omega \mid  \mathcal{L}(\theta_N(\omega)) \le \inf_{\theta \in \Theta} \mathcal{L}(\theta) + 2\frac{(\ln\dfrac{2}{\delta}+ 2C_{\text{pac}}) \ln(N)}{\sqrt{N}} \right\}
  %  \]
  % and since $\bP(A) \ge 1-2\delta$, the probability of the latter event is also at least $1-2\delta$.
\end{pf}
Finally, we derive parameter estimation bounds, for which we need the following assumption.
\begin{Assumption}
\label{assumption:strong_convexity}
There exist $\theta_{\rm true}\in\arg\min_{\theta\in\Theta}\mathcal{L}(\theta)$ and $m_\Theta>0$ such that, for every $\theta\in\Theta$,
$\mathcal{L}(\theta)-\mathcal{L}(\theta_{\rm true})\ge m_\Theta\|\theta-\theta_{\rm true}\|_2^2$. 
  % The true error map $\theta \mapsto \mathcal{L}(\theta)$ is smooth and strongly convex
  %   in $\theta$ and its Hessian is bounded from below
  %   $m_{\Theta}I \preceq \nabla^2_{\theta}\mathcal{L}(\theta)$
  %   for all $\theta \in \Theta$.
\end{Assumption}
% Let 
% \[ \theta_{\text{true}} = \argmin_{h \in \mathcal{H}} \mathcal{L}(h) \]
% which exists and is unique as $\Theta$ is compact and
% $\mathcal{L}(\theta)$ is strongly convex in $\theta$ and there exists \(\theta_{\rm true}\in\arg\min_{\theta\in\Theta}\mathcal{L}(\theta)\) and $m_\Theta>0$ such that $\mathcal{L}(\theta) - \mathcal{L}(\theta_{\text{true}}) \ge m_{\Theta} \|\theta - \theta_{\text{true}}\|_2^2$ holds for every $\theta \in \Theta$.
%This assumption implicitly encodes both structural identifiability and persistence of excitation, as it implies that 
%\begin{equation}
%  \label{eq:strong_convexity}
%\mathcal{L}(\theta) - \mathcal{L}(\theta_{\text{true}}) \ge m_{\Theta} \|\theta - \theta_{\text{true}}\|_2^2 
%\end{equation}
This assumption implies \emph{structural identifiability} around $\theta_{\text{true}}$: if $\theta \ne \theta_{\text{true}}$, the models have different input-output behavior. Otherwise, $\mathcal{L}(\theta) = \mathcal{L}(\theta_{\text{true}})$, contradicting $m_{\Theta} \|\theta - \theta_{\text{true}}\|_2^2 > 0$. 
It also ensures that $\rvw$ is \emph{persistently exciting}: if two models have identical output response to $\rvw$, then $\rvy_{\theta}(t)=\rvy_{\theta_{\text{true}}}(t)$, so $\mathcal{L}(\theta) = \mathcal{L}(\theta_{\text{true}})$, which implies $\theta = \theta_{\text{true}}$.
\begin{corollary}[Parameter estimation]
  \label{cor:parameter_estimation}
    With the assumptions of Theorem \ref{thm:oracle_inequality} and Assumption \ref{assumption:strong_convexity},
    \begin{multline}
     \bP(\|\hat \theta_N - \theta_{\text{true}}\|_2^2 \le 2\varepsilon_N(\delta)/m_{\Theta}) \ge 1-2\delta
    \end{multline}
    for all $\delta \in (0,0.5)$, where $\hat{\theta}_N \in \mathrm{argmin}_{\theta \in \Theta} \hat{\mathcal{L}}_N(\theta)$. 
\end{corollary} 
The result above bounds the parameter estimation rate at $O(\sqrt{\log N}/N^{1/4})$, which is slower than some existing rates \cite{Ziemann1,Ziemann2,racz2025finite,pmlr-v283-sattar25a}. However, it applies to a much broader class of learning problems, so this generality comes at a cost in convergence speed.

%One may wonder when the conditions on strong convexity of $\mathcal{L}(\theta)$ are satisfied.
%Autoregressive models with a linear dependence on parameters satisfy these
%conditions.
In the ARX example below, Assumption~\ref{assumption:strong_convexity} holds when the regressor covariance matrix \(Q=\mathbb E[\rvz(t)\rvz(t)^\top]\) is positive definite with \(m_\Theta=\lambda_{\min}(Q)\).
\begin{Remark}[Autoregressive models]
    \label{rem:fir:param}
    Consider the SISO case ($n_y=n_u=1$).
    Let  $\rvy$ be generated by an ARX model
    \[ \rvy(t)=\sum_{i=1}^{n} a_i \rvy(t-i) + \sum_{i=1}^{n} b_i \rvu(t-i) + \rve(t) \]
    where $\rvu$ is a bounded stationary i.i.d. process independent
    of $\rve$ and $\rve$ is also a bounded i.i.d process,
    and assume that the hypothesis class is formed as follows with $\theta \in \Theta \subseteq \mathbb{R}^{2n}$
    \[ %h_{\theta}(\{\rvw(s)\}_{s < t}) 
    \hat{\rvy}_{\theta}(t \mid t_0)= \theta^T \rvz(t), ~
       \rvz(t)=\begin{bmatrix} \rvw^T(t-1) & \cdots & \rvw^T(t-n) \end{bmatrix}^T. \]
    %If $\rve$ is bounded, 
    If we assume squared loss, $\mathbb E[e(t)]=0$, and that $y$ is the causal stationary solution of the stable ARX model and we let $\Theta$ satisfy the conditions of Theorem~\ref{thm:mainThm} and contain $\theta_{\rm true}=(a_1,b_1,\ldots,a_n,b_n)^\top$, then we can represent both the process $\rvy$ and the predictions $\hat{\rvy}_{\theta}(t\mid 0)$ as outputs of system of class \class{}. %, \cite{Arxive}. 
    It can be shown that % \cite{Arxive} that
    \[ \mathcal{L}(\theta)=(\theta-\theta_{true})^T Q (\theta-\theta_{true})
       + \sigma_{\rve}^2
    \]
    where  $\sigma_{\rve}^2=E[\rve(t)^2]$ and 
    $Q=E[\rvz(t)\rvz(t)^T]$.  %with $\rvz(t)=\begin{bmatrix} \rvy^T(t-1) & \cdots & \rvy^T(t-n) & \rvu^T(t-1) & \cdots & \rvu^T(t-n) \end{bmatrix}^T$, 
    If $Q \succ 0$, Assumption~\ref{assumption:strong_convexity} holds with $m_{\Theta}=\lambda_{\min}(Q)$, and Corollary~\ref{cor:parameter_estimation} applies to any measurable minimization of $\hat{\mathcal{L}}_N$ over $\Theta$.
    %hence $\mathcal{L}(\theta)$ is strongly convex with $m_{\Theta}=\lambda_{\min}(Q)$.
    %Then if $\hat{\theta}_N$ is the minimizer of the empirical loss
    %\( \frac{1}{N} \sum_{t=1}^{N} (\hat{\rvy}(t)-\rvy(t))^2 \), which, in this 
    %case is the linear least squares solution of the regression problem,
    %then Corollary \ref{cor:parameter_estimation}. 
  \end{Remark}
%    Moreover, it can be shown that 
%
%In many cases, convexity of the true loss is too much to ask for, if finite dimensional parametrization is used. However, for LTI and bilinear systems, we can use a parametrization in terms of the infinite sequence of Markov parameters, in which case the true error for quadratic loss function will be a quadratic function of Markov parameters, and a reasoning similar to that of Corollary \ref{cor:parameter_estimation} can be used to bound the difference between 
%Markov parameters. 
%
%
\section{Bounds for parameter estimation of (bi)linear systems using $H_2$ norm}
\subsection{Bound for $H_2$ norm for \textbf{LTI systems}}
Let
the data  $\rvy$ to be generated by an LTI system in innovation form
\begin{equation}
\begin{aligned}
  \hat{\rvs}_{g}(t+1) &= A_0\hat{\rvs}_{g}(t)+B_0\rvu(t)+K_0\rve^s(t), \\
  \rvy(t) &= C_0\hat{\rvs}_{g}(t)+\rve^s(t), \qquad t \in \sZ,
\end{aligned}
\label{eq:T0sys1}
\end{equation}
where $A_0,B_0,K_0,C_0,D_0$ are matrices of suitable dimensions, $A_0$ and $A_0-K_0C_0$ are both Schur, and $\rve^s(t)$ is a zero-mean i.i.d. bounded noise
$\|\rve^s(t)\|_2 \le C_{\rve^s}$.
Let us assume that $\rvu$ is a stationary bounded process generated by a stochastic LTI system driven by bounded noise, i.e.,
\begin{equation}\label{eq:u_gen}
    \rvs_\rvu(t\!+\!1)\!=\!F_u\rvs_\rvu(t)+H_u\rve_u(t), ~~ \rvu(t)\!=\!G_u\rvs_\rvu(t) +\rve_u(t),
\end{equation}
where $\rve_u$ is zero mean, bounded, i.i.d. , $F_u$ and $F_u-H_uG_u$ are Schur and 
$\{\rve_u(s)\}_{s \in \sZ}$ and $\{\rve^s(s)\}_{s \in \sZ}$ are independent
and $\|\rvu(t)\|_2 \le C_\rvu$.
Then
by repeating the construction  in \cite{eringis2021optimal}, 
it follows that $\rvw (t)=\begin{bmatrix} \rvy^T & \rvu^T \end{bmatrix}^T$ is generated by a class \class{} system,
obtained by series interconnection of \eqref{eq:T0sys1} and \eqref{eq:u_gen}:
%such that
%\begin{color}{black}
%\begin{equation}\label{eq:predictor_lti_input}
%  \begin{aligned}
%   \rvs(t+1) &= \underbrace{\begin{bmatrix} A_0 & B_0G_u \\0 & F_u \end{bmatrix}}_{A_g}
%    \rvs(t)+\underbrace{\begin{bmatrix}K_0&B_0\\0&H_u\end{bmatrix}}_{K_g}\rve_g(t), \\
%    \rvw(t)   &= \underbrace{\begin{bmatrix} C_0 & 0 \\ 0 & G_u \end{bmatrix}}_{C_g}\rvs(t)+\rve_g(t).
%  \end{aligned}
%\end{equation}
%\end{color}
	%where $A_g=$, $K_g=\begin{bmatrix}K_0&B_0\\0&H_u\end{bmatrix}$,
 % $C_g=\begin{bmatrix} C_0 & 0 \\ 0 & G_u \end{bmatrix}$, $\rve_g=\begin{bmatrix} \e^s \\ \rve_u \end{bmatrix}$.
  %Let us define 
  %We take the observed process $w(t)=(y(t)^\top,u(t)^\top)^\top$ to be generated by the stationary solution of \ref{eq:T0sys1} and \ref{eq:u_gen}. Then \begin{color}{black} the generator \end{color} satisfies Assumption \ref{as:generator}:
  \begin{lemma}
  \label{lemma:LTI:classS_g}
  %%  $\tau$ be the spectral radius of $A_g$ and $f_{g}(\vs,\vv)=A_g\vs+K_g\vv$ and $g_g(\vs,\vv)=C_g\vs+\vv$
  Let $n$ be the sum of state-space dimensions of \eqref{eq:u_gen} and \eqref{eq:T0sys1}.
  There  exist compact sets
  $\mathbb{S}_g \subseteq \mathbb{R}^n$, $\mathbb{V}_g, \mathbb{Y}_g \subseteq \mathbb{R}^{n_u+n_y}$ such that for
  \begin{align*}
     f_{g}(\rvs,\rvv)=A_g\rvs+K_g\rvv, \quad g_g(\rvs,\rvv)=C_g\rvs+\rvv \\
     A_g=\begin{bmatrix} A_0 & B_0G_u \\0 & F_u \end{bmatrix}, \quad K_g=\begin{bmatrix}K_0&B_0\\0&H_u\end{bmatrix},
     C_g=\begin{bmatrix} C_0 & 0 \\ 0 & G_u \end{bmatrix}
  \end{align*}
  it holds that $f_g(\mathbb{S}_g \times \mathbb{V}_g) \subseteq \mathbb{S}_g$, $g_g(\mathbb{S}_g \times \mathbb{V}_g) \subseteq \mathbb{Y}_g$, and 
  %for $f_{g}(\vs,\vv)=A_g\vs+K_g\vv$ and $g_g(\vs,\vv)=C_g\vs+\vv$, 
  the system $\mathcal{S}_g=(\mathbb{S}_g,\mathbb{V}_g,\mathbb{Y}_g,f_g|_{\mathbb{S}_g \times \mathbb{V}_g},g_g|_{\mathbb{S}_g \times \mathbb{V}_g})$ satisfies Assumption \ref{as:generator}.
  %is a system of class \class{} such that 
  %$\rve_g$ takes values in $\mathbb{V}_g$ and the  response  of $\mathcal{S}_g$ to $\rve_g$   is $\rvw$, i.e., $\rvw(t)=y_{\mathcal{S}_g}(\rve_g)$.
  \end{lemma} 
  The proof of Lemma \ref{lemma:LTI:classS_g} % is in the appendix%\cite{Arxiv}
   is based on Lemma \ref{lemma:P:classS} and presented in \cite{arxiv}.

%  $\|A_g^k\|_2 \le C_g \tau_q^k$.
%  It then follows that if $C_{n}=\|\rve_g(0)\|_{\infty} \le \|\rve_u(0)\|_{\infty}+\|\e^s(0)\|_{\infty}$, and we set 
%  $\mathbb{S}_g=\{ x \mid \|x\|_2 \le (n_y+n_u)C_n^2\|K_g\|_{2}\frac{C_g}{1-\tau_q} \}$, 
%  $\mathbb{V}_g=\{ \vv  \mid \|\vv\|_{\infty} \le C_n \}$, $\mathbb{Y}_g=\{ \vw  \mid \|\vw\|_{2} \le C_n\sqrt{n_y+n_u}(\frac{C_g}{1-\tau_q}\|C_g\|_2\|K_g\|_{2}+1) \}$, $f_g(\vs,\vv)=A_g\vs+K_g\vv$, $g_g(\vs,\vv)=C_g\vs+\vv$, then 

  Let $\Theta \subseteq \mathbb{R}^{n_{\theta}}$ be a compact set satisfying Assumption \ref{as:parameter_set}
   and
  consider the parameterization $\hat{f}(\rvs,\rvw,\theta)=\hat{A}_\theta\rvs+\hat{B}_\theta\rvw$,
  $\hat{g}(\rvs,\rvw,\theta)=\hat{C}_\theta\rvs$, such that $\hat{A}_{\theta},\hat{B}_{\theta},\hat{C}_{\theta},\hat{D}_{\theta}$
  are Lipschitz continuous in $\theta$, and $\sup_{\theta \in \Theta}\|\hat{A}_{\theta}\| < \tau <1$ %for some $\tau \in (0,1)$, 
  and $\hat{\rvs}_{s}(\theta)=0$. 
  \begin{lemma}
  \label{classS:LTI:predictor}
    There exist bounded sets $\mathbb{S},\mathbb{W},\mathbb{Y}$ such that
    %$\hat{A}_{\theta}\mathbb{S}+\hat{B}_{\theta}\mathbb{V} \subseteq \mathbb{S}$,
    %$\hat{C}_{\theta}\mathbb{S}+\mathbb{V} \subseteq \mathbb{Y}$, and
    %for $\hat{f}_{\theta}(\vs,\vv)=\hat{A}_\theta\vs+\hat{B}_\theta\vv$ and $\hat{g}_{\theta}(\vs,\vv)=\hat{C}_\theta\vs+\vv$, 
    the system $\mathcal S_{\rm param}=(\mathbb S, \mathbb W\times\Theta,\mathbb{Y},\hat f|_{\mathbb S \times \mathbb W\times\Theta},\hat g|_{\mathbb S \times \mathbb W\times\Theta})$ 
 %   $\mathcal{S}_{param}=(\mathbb{S},\mathbb{W},\mathbb{Y},\hat{f},\hat{g})$ 
    and the map $\hat{\rvs}_s$  satisfies Assumption \ref{as:parametrisation}.
    %ssuis a system of class \class{} and its input-output map induced by the initial state is the same as that of \eqref{eq:predictor:LTI}.
    %Moreover, $\rvw(t)$ takes values in $\mathbb{V}$ and
    %$h(\{\rvw(s)\}_{s=0}^{t})=y_{\mathcal{S}_{\theta}}(0,0,\rvw);t)=\hat{\rvy}_{\theta}(t\mid 0)$
    %and $\lim_{t_0 \to -\infty} h(\{\rvw(s)\}_{s=t_0}^{t})=y_{\mathcal{S}_{\theta}}(\rvw)(t)=\hat{\rvy}_{\theta}(t)$.
  \end{lemma}
  The proof relies on using Lemma  \ref{lemma:P:classS}. % and is in the appendix. %\cite{Arxiv}. 

  Consider the quadratic loss $\ell(y,y')=\|y-y'\|_2^2$. 
  Define the \emph{Markov parameters} $M_k(\theta)=\hat{C}_{\theta}\hat{A}_\theta^{k-1}\hat{B}_\theta$,
  $M(\theta)=\{M_k(\theta)\}_{k=1}^{\infty}$
  and $M_k=C_0(A_0-K_0C_0)^{k-1}\begin{bmatrix} K_0 & B_0 \end{bmatrix}$, $M=\{M_k\}_{k=1}^{\infty}$.
   It is well-known that
   %\begin{color}{black}
  %\[ 
   \begin{align*}
     \hat{\rvy}_{\theta}(t)\!=\!\sum_{k=1}^{\infty} \hat{M}_k(\theta)\rvw(t\!-\!k), \quad 
     \rvy(t)\!=\!\sum_{k=1}^{\infty} M_k\rvw(t\!-\!k) +\rve^s(t)
   \end{align*}
  %\]
  %   \end{color}
   %If we denote by $M$ the sequence $\{M(k)=C_0(A_0-K_0C_0)^k\begin{bmatrix} K_0 & B_0 \end{bmatrix}\}_{k=0}^{\infty}$, 
   %and by $\hat{M}_{\theta}=\{\hat{M}_{\theta}(k)=\hat{C}_{\theta}\hat{A}_\theta^k\hat{B}_\theta\}_{k=0}^{\infty}$ and 
  We denote by  $\|Z\|_{H_2}=\sqrt{\sum_{k=1}^{\infty}\|Z(k)\|_F^2}$ the usual $H_2$ norm of a system with an absolutely summable impulse response $Z=\{Z(k)\}_{k=1}^{\infty}$. Note that due to stability of the predictors and data generators, 
   $M,M_{\theta}$ are absolutely summable, and so is $M-M_{\theta}$.
   It is easy to see that 
  \[ 
    \begin{split}
     \mathcal{L}(\theta)=& \mathrm{tr}\sum_{i,j=1}^{\infty} (M_i-M_i(\theta))Q_{i,j}(M_j-M_j(\theta))^T+ \sigma_{\rve_s}^2   \\
     & \ge m_x \|M-M_{\theta}\|_{H_2}^2 + \sigma_{\rve_s}^2,
    \end{split}
  \]
  where $Q_{i,j}=E[\rvw(t\!-\!i)\rvw(t\!-\!j)^T]$ and $\sigma_{\rve_s}^2=E[\|\rve_s(t)\|_2^2]$, and 
  $m_x$ is the infimum of the smallest eigenvalues of the matrices $(Q_{i,j})_{i,j=1}^{N}$. 
 % Let us assume that 
 % $m_x > 0$. 
  %It then follows that
  %\[ \mathrm{trace}\sum_{i,j=0}^{\infty} (M(i)-M_{\theta}(i))Q_{i,j}(M(j)-M_{\theta}(j))^T \ge m_x \|M-M_{\theta}\|_{H_2}^2 \]

  Assume that there exists $\theta_{true} \in \Theta$ such that $M=M(\theta_{true})$.
  Then $\inf_{\theta \in \Theta} \mathcal{L}(\theta) = \sigma_{\rve_s}^2=\mathcal{L}(\theta_{true})$, and 
  $\mathcal{L}(\theta)-\inf_{\theta \in \Theta} \mathcal{L}(\theta) \ge m_x \|M-M(\theta)\|_{H_2}^2.$

  By repeating the proof of Corollary \ref{cor:parameter_estimation}, we get the following $H_2$ norm error bound for the empirical minimizer $\hat{\theta}_N \in \arg\min_{\theta\in\Theta}\widehat{L}_N(\theta)$.
  \begin{corollary}[$H_2$ parameter estimation error]
    \label{col:lti}
    Assume that $m_{x} > 0$. Then 
   \[ 
\bP\!\left(
\|M(\hat\theta_N)-M(\theta_{\rm true})\|_{H_2}^{2}
\le \frac{2\varepsilon_N(\delta)}{m_x}
\right)\ge 1-2\delta.
     %\bP(\|M_{\hat{\theta}_N}-\hat{M}_{\theta_{\text{true}}}\|_H^2 \le 
     %\le 2(\ln\dfrac{2}{\delta}+ 2C_{\text{pac}})/m_{x} \frac{\ln(N)}{\sqrt{N}}) \ge 1-2\delta
  \]
  \end{corollary}
 %Note that $m_x$ is the minimum eigenvalue of the covariance matrix of the past data and it is a direct characterization of
 %Mpersistence of excitation. 
 The assumption $m_x > 0$ means that the data is persistently exciting.
Corollary \ref{col:lti} aligns with \cite[Corollary 3.4]{Nips2026}, with constants upper-bounding those in the prior work.

It is well-known (see \cite{RalfPeeters}) that the $H_2$ norm topology on stable minimal LTI systems coincides with the Riemannian manifold topology of these equivalence classes modulo isomorphism.
Assume that for every $\theta \in \Theta$, the system $(\hat{A}_{\theta},\hat{B}_{\theta},\hat{C}_{\theta})$ is minimal and structurally identifiable: the only isomorphism $T$ between two parametrizations corresponds to $\theta_1=\theta_2$ with $T$ the identity.
Then the map $\iota:\Theta \ni \theta \mapsto$ [equivalence class of $(\hat{A}_{\theta},\hat{B}_{\theta},\hat{C}_{\theta})$] is injective into the manifold of stable minimal systems. By this identifiability, we may define the metric $d(\theta_1,\theta_2) \coloneqq \|M(\theta_1)-M(\theta_2)\|_{H_2}$ on $\Theta$.
Thus Corollary \ref{col:lti} provides a parameter estimation error bound in the $H_2$ distance on impulse responses.
 
%\paragraph*{Bound for $H_2$ norm for \textbf{bilinear systems}}
\subsection{Bound for $H_2$ norm for \textbf{bilinear systems}}
  %We can extend the analysis above to bilinear models as follows. 
  Assume that the input
  $\rvu$ is an i.i.d. process with zero mean and uncorrelated components, 
  such that $\|\rvu(t)\|_{\infty} \le 1$ and assume that $\rvy$ is a stationary square integrable 
  process generated by a bilinear system
  \begin{equation}
    \label{eq:T0sys_bilinear}
    \begin{aligned}
		&\hat{\rvs}_{g}(t+1)=A(\rvu(t))\hat{\rvs}_g(t)+K(\rvu(t))\rve^s(t) \\
    %A_0+\sum_{i=1}^{n_u}A_i\rvu_i(t))\hat{\rvs}_{g}(t)+{} \\
		%&(K_0,\sum_{i=1}^{n_u} K_i\rvu_i(t))\e^s(t), \quad
    & \rvy(t) = C_0\hat{\rvs}_{g}(t)+\rve^s(t), \quad t \in \sZ, \\
    & A(u)=A_0+\sum_{i=1}^{n_u}A_iu_i, \quad K(u)=K_0+\sum_{i=1}^{n_u} K_i u_i,
    \end{aligned}
  \end{equation}
  where 
 % with the convention $\rvu_0(t) \equiv 1$, so that $A(\rvu(t))=\sum_{i=0}^{n_u} A_i\rvu_i(t)$ and  $K(\rvu(t))=\sum_{i=0}^{n_u} K_i\rvu_i(t)$
  and
  we assume that \textbf{(1)} $(1+n_u)\sum_{i=0}^{n_u} (A_i \otimes A_i)$ is Schur with a maximal eigenvalue $\tau \in (0,1)$
  and \textbf{(2)} $(1+n_u)\sum_{i=0}^{n_u} (A_i-K_iC_0) \otimes (A_i-K_iC_0)$ is Schur with a maximal eigenvalue $\hat{\tau} \in (0,1)$
  and \textbf{(3)} $\{\e^s(t)\}_{t \in \sZ}$ is a zero mean i.i.d. bounded  process and 
  $\{\rvu(t)\}_{t \in \sZ}$ and $\{\e^s(t)\}_{t \in \sZ}$ are independent.
  It can be shown that \eqref{eq:T0sys_bilinear} has a wide-sense stationary solution based 
  on \cite[Lemma 1]{TAC:MPRV}.
  %, see appendix. %\cite{Arxiv}. 
  %In fact, we will show that $\rvy$ is stationary
  %as \eqref{eq:T0sys_bilinear} can be viewed as a system of class \class{}. 
  %The key observation is the counterpart of Remark \ref{rem::contract}.\\
 % The convergence of the right-hand sides of \eqref{eq:bilinear_output} are understood in the 
 %We call $M(w)=C_0 A_w \begin{bmatrix} K_0 & B_0 \end{bmatrix}$ the \emph{Markov parameter} corresponding to the sequence $w$ and 
 %we call $M=\{M_w\}$ the Markov parameters of \eqref{eq:T0sys_bilinear}.
 %Notice that \eqref{eq:T0sys_bilinear} can be rewritten as a class \class{} system.
 This generator satisfies Assumption \ref{as:generator}:
 \begin{lemma}
  \label{ge:systemS:bilinear}
   \begin{color}{black}There exist a  compact \end{color} set 
   \( \mathbb{S}_g \subseteq \mathbb{R}^{n} \) and bounded sets \( \mathbb{V}_e \subseteq \mathbb{R}^{n_y},
     %\mathbb{V}_g=[-1,1]^{n_u} \times \mathbb{V}_e, 
     \mathbb{Y}_g \subseteq \mathbb{R}^{n_y+n_u} \)
   such that %$\mathbb{S}_g$ is compact and 
   the functions 
   \begin{align*}
    & f_g: \mathbb{S}_g \times [-1,1]^{n_u} \times \mathbb{V}_e \ni {\color{black}(\vs,\vu,\ve)} \mapsto A(\vu)\vs+K(\vu)\ve,  \\
    & g_g: \mathbb{S}_g \times [-1,1]^{n_u} \times \mathbb{V}_e \ni (\vs,\vu,\ve) \mapsto \begin{bmatrix} C_0\vs+\ve \\ \vu \end{bmatrix} 
  \end{align*}
 take values in $\mathbb{S}_g$ and $\mathbb{Y}_g$ respectively, 
   and 
   \[\mathcal{S}_g=(\mathbb{S}_g,\mathbb{V}_g=[-1,1]^{n_u} \times \mathbb{V}_e,\mathbb{Y}_g,f_g,{\color{black}g_g})\] 
   is a class \class{} system, and
  %\[ [\rvu^T,\rve_s^T]^T \in \mathbb{V}_g, \quad \rvw=[\rvy^T, \rvu^T]^T \in \mathbb{Y}_g, \]
   and
    \[
    \begin{bmatrix}\rvy^T, & \rvu^T \end{bmatrix}^T={\color{black}\rvy_{\mathcal{S}_g}([\rvu^T,\rve_s^T]^T)}
    \]
   % In particular, $\rvw$ is stationary.
   %for all $\vs  \in \mathbb{S}_g$, $\vu \in [-1,1]^{n_u}$, $\vv \in \mathbb{V}_e$, $\vu \in \mathbb{R}^{n_u}$.
   %$f_g(\mathbb{S}_g,\mathbv{V}_g) \subseteq \mathbb{S}_g$
 \end{lemma}
 The proof of Lemma \ref{ge:systemS:bilinear}% is in the appendix%\cite{Arxiv}
 %, it 
  is based on Lemma \ref{lemma:P:classS}, se \cite{arxiv}.
Let $\Theta \subseteq \mathbb{R}^{n_{\theta}}$ be a set satisfying Assumption \ref{as:parameter_set} and consider the following predictors
%which can also be viewed as bilinear or LPV (Linear Parameter-Varying) systems %with $\hat{\rvs}(t_0) = 0$:
\begin{equation}
\label{pred:bilinear}
\begin{aligned}
&\hat{\rvs}(t+1 \!\!\mid\!\! t_0) \!=\!  \hat{A}^{[\theta]}(\rvu(t))\hat{\rvs}(t \!\!\mid\!\! t_0)+ \hat{K}^{[\theta]}(\rvu(t))\rvy(t) \\
& \hat{\rvy}_{\theta}(t \!\!\mid\!\! t_0) \!=\! \hat{C}(\theta)\hat{\rvs}(t \!\!\mid\!\! t_0) \quad
   \hat{\rvs}(t_0 \!\!\mid\!\! t_0)=0, \\
& \hat{A}^{[\theta]}(u)\!\!=\! \!\!\hat{A}_{0}^{[\theta]}\!+\!\sum_{i=1}^{n_u}\! \hat{A}_{i}^{[\theta])}u_i, ~~ \hat{K}^{[\theta]}(u)\!\!=\!\!\hat{K}_{0}^{[\theta]}\!+\!\sum_{i=1}^{n_u}\! \hat{K}_{i}^{[\theta]}u_i 
\end{aligned}
\end{equation}
Assume that \textbf{(1)} \(\{\hat A_i^{[\theta]},\hat K_i^{[\theta]}\}_{i=0}^{n_u}\), and \(\hat C^{[\theta]}\) are Lipschitz continuous in \(\theta\) on \(\Theta\), \textbf{(2)} there exist \(P\succ0\) and \(0<\hat\tau<1\) such that, for every \(\theta\in\Theta\),
\(
(1+n_u)\sum_{i=0}^{n_u}\hat A_i^{[\theta]^\top} P\hat A_i^{[\theta]}
\preceq \hat\tau^2P
\)
and
\textbf{(3)} $\|{\color{black}\hat{C}(\theta)}\|_2 \le \hat{L}_{g,1}$, $\sum_{i=0}^{n_u}\|\hat{K}_{i}(\theta)\|_2 \le \hat{L}$ \begin{color}{black} for all $\theta \in \Theta$\end{color}.
%where the maps \(\theta\mapsto\hat A_i(\theta),\hat K_i(\theta)\), \(i=0,\ldots,n_u\), and \(\theta\mapsto\hat C(\theta)\) are Lipschitz on \(\Theta\). Assume that there exist a matrix \(P\succ0\) and a constant \(0<\hat\tau<1\) such that, for every \(\theta\in\Theta\),
%\[
%(1+n_u)\sum_{i=0}^{n_u}\hat A_i(\theta)^\top P\hat A_i(\theta)
%\preceq \hat\tau^2P.
%\]

%where $\hat{A}(\theta,\rvu)=\hat{A}_{0}(\theta)+\sum_{i=1}^{n_u} \hat{A}_{i}(\theta){\color{black}\rvu_i(t)}$ and $\hat{K}(\theta,\rvu)=\hat{K}_{0}(\theta)+\sum_{i=1}^{n_u} \hat{K}_{i}(\theta){\color{black}\rvu_i(t)}$, \begin{color}{black}the matrices $\hat{A}_i(\theta),\hat{K}_i(\theta),\hat{C}(\theta)$ are continuous in $\theta$,\end{color} and 
%\begin{color}{black}the spectral radius of \end{color}$(1+n_u)\sum_{i=0}^{n_u} (\hat{A}_{i}(\theta) \otimes \hat{A}_{i}(\theta))$ is \begin{color}{black}smaller than $\hat{\tau}^2$\end{color},

The predictor can be viewed as an LPV \begin{color}{black}system\end{color} with affine dependence on scheduling parameter, if $\rvu$ is viewed as a scheduling signal.
Alternatively, we can view \eqref{pred:bilinear} as a candidate noisy bilinear system which generates $\rvy$, if we define
the noise process $\hat{\rve}_{\theta}(t \mid t_0)=\rvy(t)-\hat{\rvy}_{\theta}(t\mid t_0)$; in this case \eqref{pred:bilinear} becomes with $\hat{\rvs}(t_0) = 0$ and 
\begin{equation*}
%\label{pred:bilinear:gen}
\begin{aligned}
  & \hat{\rvs}(t+1)= A^{[\theta]}(\rvu(t))\hat{\rvs}(t)+ \hat{K}^{[\theta]}(\rvu(t))\hat{\rve}_{\theta}(t \mid t_0), \\
  & A^{[\theta]}(u)=\hat{A}^{[\theta]}(u)+\hat{K}^{[\theta]}(u)\hat{C}^{[\theta]} \\
  &  \rvy(t) = \hat{C}^{[\theta]}\hat{\rvs}(t)+\hat{\rve}_{\theta }(t \mid t_0)
\end{aligned}
\end{equation*}
The process $\hat{\rve}_{\theta}(t \mid t_0)$ represents the prediction error of the noisy bilinear system. In general, it is not an i.i.d. process. At the true parameter, the stationary prediction error equals the i.i.d. innovation $e_s(t)$.
%However, under mild assumptions, the optimal predictor can be viewed as an i.i.d. process, as we will see below. 

We first show that the predictors \eqref{pred:bilinear} can be viewed as 
class \class{} systems satisfying Assumption \ref{as:parametrisation}.
\begin{lemma}
  \label{ge:classS:bilinear:predictor}
  There exists a compact $\mathbb{S} \subseteq \mathbb{R}^{n_{\hat s}}$ where $n_{\hat s}=\dim\hat s(t)$, and 
  bounded sets $\mathbb{V}_y \subseteq \mathbb{R}^{n_y}$, 
  $\mathbb{Y} \subseteq \mathbb{R}^{n_y}$ such that for $\mathbb{V}=[-1,1]^{n_u} \times \mathbb{V}_y$, 
  the functions 
  \begin{align*}
  & \hat{f}_{\theta}:\mathbb{S} \times \mathbb{V} \ni (\vs,\vu,\vv)
  \mapsto \hat{A}^{[\theta]}(\vu)\vs+\hat{K}^{[\theta]}(\vu)\vv, \\
  & \hat{g}_{\theta}:\mathbb{S} \times \mathbb{V} \ni (\vs,\vu,\vv)
  \mapsto \hat{C}^{[\theta]}\vs
  \end{align*}
  are well-defined, take values in $\mathbb{S}$ and $\mathbb{Y}$, respectively,
  and $\mathcal{S}_{\theta}=(\mathbb{S},\mathbb{V},\mathbb{Y},\hat{f}_{\theta},\hat{g}_{\theta})$
  is class \class{} system, and with $\hat{\rvs_s}(\theta)=0$ satisfies Assumption \ref{as:parametrisation}. 
\end{lemma}

\begin{color}{black}The proof depends on Lemma \ref{lemma:P:classS}, \cite{arxiv}.
  % and on the argument that appears in the proof of Lemma \ref{ge:systemS:bilinear}, which can be seen in the appendix. % \cite{Arxiv}.
Thus the data generator \eqref{eq:T0sys_bilinear} %leads to a class \class{}
%system 
satisfies Assumption \ref{as:generator} and the predictor \eqref{pred:bilinear}
satisfies Assumption \ref{as:parametrisation}.\end{color}

Next, we present Markov parameters and the $H_2$ norm for bilinear systems.
Let $\Sigma=\{0,\ldots,n_u\}$, $\Sigma^{*}$ the set of all finite sequences over $\Sigma$, $\epsilon$ the empty sequence, and $|w|$ the length of $w \in \Sigma^*$ (with $|\epsilon|=0$). Denote $\Sigma^{+}=\Sigma^{*} \setminus \{\epsilon\}$ and $\Sigma^k$ the set of sequences of length $k$. 
For matrices $\{A_i\}_{i \in \Sigma}$ and $\sigma=\sigma_1\ldots\sigma_k \in \Sigma^*$, define 
\[A_{\sigma}=A_{\sigma_k}\cdots A_{\sigma_1} \text{ and } A_{\epsilon}=I.\]

  For any process $\rvr$, define $\rvz_w^{\rvr}(t)$ recursively as follows: 
  \[\rvz^{\rvr}_{\epsilon}(t)=\rvr(t), \quad \rvz^{\rvr}_{wq}(t)=\rvz^{\rvr}_{w}(t-1)\rvu_q(t-1), ~ w \in \Sigma^*, q \in \Sigma.\]
  Let us define the Markov-parameters of \eqref{eq:T0sys_bilinear}
 $M,M_\theta:\Sigma^+\to\mathbb R^{n_y\times n_y}$
  %$M, M_{\theta}:\Sigma^{+} \rightarrow \mathbb{R}^{n_u+n_y }$ and $M_{\theta}:\Sigma^{+} \rightarrow \mathbb{R}^{n_u+n_y }$
  as follows: for all $w \in \Sigma^{*}$, $q \in \Sigma$,
  \[ 
    \begin{split}
  & M(qw)=C\bar{A}_wK_q, \text{ where } \bar{A}_q=A_q-K_qC, \\
  & M_{\theta}(qw)=\hat{C}^{[\theta]}\hat{A}^{[\theta]}_{w}\hat{K}_{q}^{[\theta]}.
  \end{split}
  \]

  Note that $M$, $M_{\theta}$ correspond to recognizable formal power series \cite{Fliess,Son:Real},
  and they uniquely determine the input-output behavior of the predictor \eqref{pred:bilinear} \cite{Petreczky2016}. 
  For a map $S:\Sigma^{+} \rightarrow \mathbb{R}^{n_y \times n_y}$ define the $H_2$ norm as 
  $\|S\|_{H_2}^2 = \sum_{k=1}^{\infty} \sum_{w \in \Sigma^{k}} \|S(w)\|_F^2$, if the latter sum exists.
  For a suitable subclass of absolutely square summable Markov-parameters the $H_2$ norm is a proper distance
  which can be computed from an LPV/bilinear state-space representation corresponding to the Markov-parameters
  and extends the classical $H_2$ norm for LTI systems \cite{BennerH2norm,MTNSSpace}.

%In particular, the series $M$ uniquely describes the input-output behavior of the predictor 
%Assume that there exists $\theta_{true} \in \Theta$ such that 
%$A_{i}(\theta_{true}) = A_i-K_iC$, $K_{i}(\theta_{true}) = K_i$, $C_{0}(\theta_{true}) = C_0$ for all $i$.
%It then follows from \cite{ElieCDC2022} that $\e^s$ is the 
  
Let $\ell$ be the quadratic loss, i.e., $\ell(y,y')=\|y-y'\|^2_2$. %It then follows from \eqref{bil:pred_error} that
The outputs and true error can be expressed via Markov parameters, which have a well-defined $H_2$ norm. 
%and hence it is easy to see that based on \cite{ElisCDC2022} that
\begin{lemma}[Markov parameters \& true loss]
  \label{lemma:bilinear:markov1}
With the notation above,   %the following holds:
%\textbf{$H_2$ norm} 
the norms $\|M_{\theta}\|_{H_2}$, $\|M\|_{H_2}$, $\|M-M_{\theta}\|_{H_2}$ are well-defined, and
%\textbf{Outputs via Markov parameters}
\[ 
  \begin{split}
   & \hat{\rvy}_{\theta}(t)=\sum_{k=0}^{\infty} \left(\sum_{w \in \Sigma^{k}}
   M_{\theta}(w) \rvz^{\rvy}_{w}(t)\right), \\
   & \rvy(t)=\sum_{k=0}^{\infty} \left(\sum_{w \in \Sigma^{k}}
   M(w) \rvz^{\rvy}_{w}(t)\right) + \e^s(t).
  \end{split}
\]
where both converge in the mean-square sense, %\(L^2\) when truncated by word length, 
and
%where the infinite sum  above converges absolutely a.s.
%and in fact also in expectation, by dominated convergence theorem and boundedness of $\rvy$
%\textbf{True loss and $H_2$ norm}
\begin{color}{black}
\begin{equation}
\label{bil:pred_error_1}
\begin{aligned}
  %&\rvy(t)-\hat{\rvy}_{\theta}(t)=\e^s(t)+ \sum_{k=1}^{\infty}\sum_{w \in \Sigma^{+},|w|=k} (M(w)-M_{\theta}(w))\z^{\rvy}_w(t) \\
 & \mathcal{L}(\theta)=\sigma^{2}_{\e^s} 
+  \\
& \mathrm{trace} \sum_{k=0}^{\infty} \sum_{\substack{v,w \in \Sigma^{k}}}
 (M-M_{\theta})(w)T_{w,v}^{\rvy}(M-M_{\theta})^T(v)  \\
 & \ge m_{\rvy} \|M-M_{\theta}\|_{H_2}^2 + \sigma^2_{\e^s}
\end{aligned}
\end{equation}
where $\sigma^{2}_{\e^s}=E[\|\e^s(t)\|^2_2]$, $T_{w,v}^{\rvy}=E[\rvz^{\rvy}_{w}(t)(\rvz^{\rvy}_{v}(t))^T]$ and $m_{\rvy}$ is such that $(T_{v,w}^{\rvy})_{v,w \in \Sigma^{k}, k \le N} > m_{\rvy}I$ for all $N \ge N_0$  for some $N_0 \in \mathbb{N}$. 
\end{color}
\end{lemma}

The proof applies the excitation bound to the Gram matrix of the regressors indexed by nonempty words of length at most \(K\), then lets $K\to\infty$.

%The proof is based on the observation from \cite{TAC:MPRV} that $E[\rvz_{qw}^{\rvy}(t)(\rvz_{q'v}^{\rvy}(t))^T]=0$ if $qw \ne q'v$ and $E[\rvz_{qw}^{\rvy}(t)(\rvz_{qw}^{\rvy}(t))^T]=\Lambda_q$, and it is in the appendix.% \cite{Arxive}

Assume that there exists $\theta_{true} \in \Theta$ such that 
$A_{i}^{[\theta_{true}]} = A_i-K_iC$, $K_{i}^{[\theta_{true}]} = K_i$, $C_{0}^{[\theta_{true}]} = C_0$ for all $i$.
$M_{\theta_{true}}=M$ and $\mathcal{L}(\theta_{true})=\sigma^2_{\e^s}$. 
In fact, from \cite{ElieCDC2022} it follows that $\hat{\rvy}_{\theta_{true}}(t)$ 
 is the minimal variance predictor of $\rvy(t)$ based on 
$\{\rvz_{w}^{\rvy}(t)\}_{w \in \Sigma^{+}}$ and 
the prediction error $\e^s(t)=\rvy(t)-C_0\rvs(t)$ is i.i.d.
%\begin{align*}
%  &\rvy(t)=\e^{s}(t)+\hat{\rvy}_{\theta_{true}}(t)=\e^s(t)\\
%  &+\sum_{k=0}^{\infty} \sum_{q \in \Sigma, w \in \Sigma^{*},|w|=k}
%   \hat{C}(\theta_{true}) \hat{A}(\theta_{true})_{w} \hat{K}_{q}(\theta_{true}) \rvz^{\rvy}_{w}(t)
%\end{align*}
%
%and hence
%In particular, the noise $\lim_{t_0 \rightarrow -\infty}\hat{\rve}_{\theta}(t \mid t_0)=\rve^s(t)$ is i.i.d., i.e., 
%the interpretation of the predictor corresponding to $\theta_{true}$ as a generator of $\rvy$ results in the data generator, 
%as the time from which $t-t_0$ for which the predictor is run converges to infinity. 
This then implies that $\inf_{\theta \in \Theta} \mathcal{L}(\theta)=\mathcal{L}(\theta_{true})=\sigma^2_{\e^s}$.

We can then state the following finite-sample bound on the parameter estimation error in terms of $H_2$ distance:
let $\hat{\theta}_N$ be the minimizer of the empirical loss $\hat{\mathcal{L}}_N(\theta)$.
\begin{corollary}[$H_2$ parameter estimation error bound]
\leavevmode\\
  Assume $m_{\rvy} > 0$, then  % that $\Theta$ satisfies the volume and small-ball conditions of Theorem~\ref{thm:mainThm} then %with the notation and assumptions above, 
\label{cor:bilin_H2_param_est_error}
  \[ \bP\left ( \|M_{\hat{\theta}_N}-M_{\theta_{true}}\|_{H_2}^2 \le
    \frac{2\varepsilon_N(\delta)}{m_\rvy}\right) \ge 1-2\delta
   \]
\end{corollary}
Note that $m_{\rvy} >0$ is a persistence of excitation condition, ensuring no linear dependence among elements of $\rvy$. For each word length $K$, we have $m_{y,K}:=\lambda_{\min}(G_K)>0$ where $G_K=(T^y_{v,w})_{v,w\in\Sigma^k,\, k \le K}$; the corollary assumes $\inf_{K\ge1}m_{y,K}>0$.

By \cite{Petreczky2016,MTNSSpace}, stable minimal bilinear/LPV systems form equivalence classes whose topology is induced by the $H_2$ norm on Markov-parameter differences. 
If the parameterization \eqref{pred:bilinear} is structurally identifiable and minimal \cite{Petreczky2016, Alkhoury2017a}, then $\Theta \ni \theta \mapsto$ equivalence class is injective. Thus $d(\theta_1,\theta_2) \triangleq \|M_{\theta_1}-M_{\theta_2}\|_{H_2}$ defines a metric on $\Theta$, and Corollary \ref{cor:bilin_H2_param_est_error} gives a finite-sample bound on $d(\hat{\theta}_N,\theta_{\text{true}})$.

\section{Conclusions}
We presented a PAC bound on the generalization error for learning nonlinear state-space models from single trajectories, leading to oracle bounds and finite-sample parameter estimation error bounds applicable to prediction error minimization. We demonstrated the approach for linear and bilinear systems using $H_2$ norm distance on Markov parameters. Future work includes tightening convergence rates, relaxing boundedness requirements, and applications to data-driven control.

\bibliography{bib.bib}
%\end{document}

\appendix
\section{Technical appendix}
\begin{Remark}{Autoregression}
Then the data generator is a linear system, i.e.,
\(\begin{bmatrix} 
0 & 1 & 0 & \dots & 0 \\  
0 & 0 & 1 & \dots & 0 \\  
\vdots & \vdots & \vdots & \ddots & \vdots \\  
0 & 0 & 0 & \dots & 1 \\  
a_n & a_{n-1} & a_{n-2} & \dots & a_1 
\end{bmatrix}_{A_g}
\vs
+ 
\begin{bmatrix} 
0 & 0\\ 
0 & 0\\ 
\vdots & \vdots\\ 
0 & 0\\ 
1 & 0
\end{bmatrix} 
\vv(t) \), 
\(g_g(\vs,\vv)= 
\begin{bmatrix} 
b_n & b_{n-1} & b_{n-2} & \dots & b_1 
\end{bmatrix}\vs+\begin{bmatrix} 0 & 1 \end{bmatrix}\vv \),
$\rve_g(t)=\begin{bmatrix} \rvu^T(t) & \rve(t) \end{bmatrix}$.

Let the hypothesis class be the class of autoregressive models of the form
 \[ 
      \hat{\rvy}(t)=\sum_{i=1}^{n} \alpha_i \rvy(t-i) + \sum_{j=1}^{n} \beta_j \rvu(t-j)
\]
i.e., $\Theta=\mathbb{R}^{2n}$, and 
    for any $\theta=(\theta_1,\ldots, \theta_{2n})^T \in \Theta$, let
    $D_{\theta}=\theta_1$,
    \[ 
    A=\begin{bmatrix} 0 & 0 & \cdots & 0 & 0 \\ 
        1 & 0 & \cdots & 0 &0 \\
        \vdots & \vdots & \cdots & \vdots & \vdots \\
        0 & 0 & \cdots & 1 & 0
    \end{bmatrix}, ~ 
    B=\begin{bmatrix} 1 \\ 0 \\ \vdots \\ 0 \end{bmatrix}, ~
    C_\theta=\begin{bmatrix} \theta_2 \\ \vdots \\ \theta_{2n} \\ 1 \end{bmatrix}^T
    \]
    Let $\theta_{true}=(a_1, \ldots, a_n, b_1, \ldots, b_n)$
    Let $\ell(y,y')=(y-y')^2$. Then 
    \[ \mathcal{L}(\theta)=(\theta-\theta_{true})^T Q (\theta-\theta_{true})
       + \sigma_{\rve}^2
    \]
    where  $\sigma_{\rve}^2=E[\rve(t)^2]$ and 
    $Q=E[\begin{bmatrix} \rvu(t-1) \\ \vdots \\ \rvu(t-n) \\ \rvy(t-1) \\ \vdots \\ \rvy(t-n) \end{bmatrix}
    \begin{bmatrix} \rvu(t-1) \\ \vdots \\ \rvu(t-n) \\ \rvy(t-1) \\ \vdots \\ \rvy(t-n) \end{bmatrix}^T]$.
    It then follows that 
    $\theta_{\text{true}}$ is the unique minimizer of $\mathcal{L}(\theta)$ and
    \[ \mathcal{L}(\theta)-\mathcal{L}(\theta_{\text{true}})=(\theta-\theta_{true})^T Q (\theta-\theta_{true}) \ge \lambda_{\min}(Q) \|\theta-\theta_{true}\|_2^2
    \]
    where $\lambda_{\min}(Q)$ is the minimum eigenvalue of $Q$.
    Then if $\hat{\theta}_N$ is the minimizer of the empirical loss
    \( \frac{1}{N} \sum_{t=1}^{N} (\hat{\rvy}(t)-\rvy(t))^2 \), which, in this 
    case is the linear least squares solution of the regression problem,
    then Corollary \ref{cor:parameter_estimation}. 
  \end{Remark}
  
\begin{color}{black}
\begin{pf}[Proof of Lemma~\ref{lemma:oldNewS}]
Let $\vs_{\vv_1}$ and $\vs_{\vv_2}$ be the stationary solutions corresponding to the bounded inputs $\vv_1$ and $\vv_2$. Since we have bounded inputs, the stationary solutions are bounded as well, i.e., there exists $M<\infty$ such that $\sup_{t\in\sZ}\|\vs_{\vv_i}(t)\|_2 \le M$, $i=1,2$.
For any $t_0 \le t$, $\vs_{\vv_i}(t)$ is the same as the solution started at time $t_0$ from the initial state $\vs_{\vv_i}(t_0)$, i.e., $\vs_{\vv_i}(t) = \vs_{\vv_i}(t \mid t_0, \vs_{\vv_i}(t_0))$.
Hence, by the UEIIS property: $\|\vs_{\vv_1}(t)-\vs_{\vv_2}(t)\|_2 \le C\tau^{t-t_0}\|\vs_{\vv_1}(t_0)-\vs_{\vv_2}(t_0)\|_2 + L\sum_{k=1}^{t-t_0}\tau^{k-1}\|\vv_1(t-k)-\vv_2(t-k)\|_2$, which can be upper bounded by $2CM\tau^{t-t_0} + L\sum_{k=1}^{\infty}\tau^{k-1}\|\vv_1(t-k)-\vv_2(t-k)\|_2$.\\
Here the sum is finite, as the inputs are bounded and $\tau<1$. Since $\tau<1$, letting $t_0\to-\infty$ the first term vanishes, which yields the inequality. %\eqref{def:classDef:mem}.
\end{pf}

\begin{pf}[Proof of Lemma~\ref{lemma:NewSI}]
We define
\setlength{\abovedisplayskip}{4pt}%
\setlength{\belowdisplayskip}{4pt}%
\[
\left\{
\begin{aligned}
  \rvs_{\theta}(t+1 \mid t_0) &= \hat{f}\bigl(\rvs_{\theta}(t \mid t_0),\rvw(t),\rvu(t),\theta\bigr),\\
  \rvs_{\theta}(t_0 \mid t_0) &= \hat{\vs}_s(\theta),\\
  \rvy_{\theta}(t \mid t_0) &= \hat{h}\bigl(\rvs_{\theta}(t \mid t_0),\rvu(t),\theta\bigr).
\end{aligned}
\right.
\]
By the UEIIS property: $\|\rvs_{\theta_1}(t \mid t_0)- \rvs_{\theta_2}(t \mid t_0)\|_{2} \le C \tau^{t-t_0} \|\hat{\vs}_s(\theta_1)- \hat{\vs}_s(\theta_2)\|_{2} + \sum_{k=1}^{t-t_0}L\tau^{k-1}\|\theta_1 - \theta_2\|_{2}$, which is further bounded by $L_{\rs} C \tau^{t-t_0} \|\theta_1-\theta_2\|_{2} + L \sum_{k=1}^{t-t_0}\tau^{k-1}\|\theta_1 - \theta_2\|_{2}$, and in the end, as this is a convex combination of $L_{\rs}C$ and $\frac{L}{1-\tau}$ times $\|\theta_1-\theta_2\|_{2}$, it can be upper bounded by $\max\{L_{\rs} C, \frac{L}{1- \tau}\} \|\theta_1-\theta_2\|_{2}$.
\\ Hence, for the output $\| \rvy_{\theta_1}(t \mid t_0)- \rvy_{\theta_2}(t \mid t_0) \|_{2} = \| \hat{h}\bigl(\rvs_{\theta_1}(t \mid t_0),\rvu(t),\theta_1\bigr) - \hat{h}\bigl(\rvs_{\theta_2}(t \mid t_0),\rvu(t),\theta_2\bigr) \|_{2}$ can be upper bounded by $L_{g,1} \|\rvs_{\theta_1}(t \mid t_0)- \rvs_{\theta_2}(t \mid t_0)\|_{2} + L_{g,2}  \|\theta_1-\theta_2\|_{2} \le \bigl(L_{g,1}\max\{L_{\rs} C, \frac{L}{1- \tau}\}+L_{g,2}\bigr)\|\theta_1- \theta_2\|_{2}$.\\
Furthermore, if $\rvs_{\theta, \rvw}(t)$ is a stationary solution of the system, $ \|\rvs_{\theta_1, \rvw}(t)- \rvs_{\theta_2,\rvw}(t)\|_{2} \le L \sum_{k=1}^{\infty} \tau^{k-1}\|\theta_1 - \theta_2\|_{2} = \frac{L}{1-\tau}\|\theta_1 - \theta_2\|_{2} $.\\
Putting the parts together, as $\rvy_{\theta}(t) = \hat{h}\bigl(\rvs_{\theta,\rvw }(t),\rvu(t),\theta\bigr)$, we get:
\[
\| \rvy_{\theta_1}(t) - \rvy_{\theta_2}(t) \|_{2}
\le \Bigl( L_{g,1} \frac{L}{1-\tau} + L_{g,2} \Bigr) \|\theta_1 - \theta_2\|_{2}.
\]
So, we can take $L_{\mathrm{pred}}$ to be $ L_{g,1} \max\{L_{\rs} C, \frac{L}{1-\tau}\} + L_{g,2} $, which covers both cases.
\end{pf}
\end{color}

\begin{pf}[Proof of Lemma~\ref{lemma:P:classS}]

By the Jacobian in \eqref{P:classS} and the notion of  $\|x\|_P=(x^\top P x)^{1/2}$ 
we have $\| J_s f(s,v)d\|_P\le\tau\|d\|_P,\forall d$ therefore for states $s_1,s_2$ and $v_1,v_2$

\[
\begin{aligned}
&\|f(s_1,v_1)-f(s_2,v_2)\|_P\\
&= \|f(s_1,v_1)-f(s_2,v_1) + f(s_2,v_1) -f(s_2,v_2)\|_P \\
&\le \|f(s_1,v_1)-f(s_2,v_1)\|_P + \|f(s_2,v_1) -f(s_2,v_2)\|_P \\
&\le \tau\|s_1-s_2\|_P +L\|v_1-v_2\|_2 .
\end{aligned}
\]
 
where $L=\sqrt{\lambda_{\max}(P)}\,L_f$. As $f(0,0)=0$ $\mathbb S=\{s:\|s\|_P\le C_s = LC_u/(1-\tau) \}$ and $\mathbb V=\{v:\|v\|_2\le C_u\}$ by the previous argument 

\[
\begin{aligned}
\|f(s,v)\|_P &\le \tau \|s\|_P + L \|v\|_2 \le \tau \|s\|_P + L C_u \\
&\le \tau LC_u/(1-\tau) + L C_u = C_s
\end{aligned}
\]

thus $f(\mathbb{S}\times \mathbb{V}) \subseteq \mathbb{S}$. Similarly, 

\[
\begin{aligned}
\|g(s,v)\|_2 &\le L_{g,1} \|s\|_2 + L_{g,2} \|v\|_2 \\
&\le L_{g,1} L C_u/((1-\tau)\sqrt{\lambda_{min}(P)}) + L_{g,2} C_u
\end{aligned}
\]

therefore $g(\mathbb{S},\mathbb{V}) \subseteq \mathbb{Y}$. 

Let us turn to UEIIS. For any two state trajectories, $s_i(t)=s_S(s_{0,i},t_0,v_i;t),i=1,2$, let us denote $\Delta s(t)=s_1(t)-s_2(t)$
and $\Delta v(t)=v_1(t)-v_2(t)$ and by repeatedly applying $\|\Delta s(t+1)\|_P
\le \tau\|\Delta s(t)\|_P+L\|\Delta v(t)\|_2$ we have 

\[
\begin{aligned}
&\|\Delta s(t)\|_2 \le C\tau^{t-t_0}\|\Delta s(t_0)\|_2\\
&+\frac{L}{\sqrt{\lambda_{min}(P)}}\sum_{k=1}^{t-t_0}\tau^{k-1}\|\Delta v(t-k)\|_2.
\end{aligned}
\]
where $C=\sqrt{\lambda_{max}/\lambda_{min}}$ via the norm inequality.

Lastly, we show convergence and uniqueness. We consider the same input and denote two trajectories with different 
starting zero time states, $s_1(t_{0,1}) =0$ and $s_2(t_{0,2})=0$ with $t_{0,1}<t_{0,2}<t$ thus 

\[
\begin{aligned}
\|s_1(t) - s_2(t)\|_P \leq \tau^{t-t_{0,2}}\|s_1(t_{0,2})\|_P \leq \tau^{t-t_{0,2}} C_s.
\end{aligned}
\]   

%thus for any $\epsilon$ there exists a $T \geq t$ that for $\|s_1(T) - s_2(T)\|_P \leq \epsilon$ and by the properties of $\mathbb S$ it has a limit if the system starts from $-\infty$ and it is recursive given $f$ is continuous. 
For fixed \(t\) and \(t_{0,1}<t_{0,2}<t\), the bound \(C_s\tau^{t-t_{0,2}}\) tends to zero as \(t_{0,2}\to-\infty\), uniformly over \(t_{0,1}<t_{0,2}\). Hence \(s_S(0,t_0,v;t)\) is Cauchy as \(t_0\to-\infty\) and has a limit \(s_v(t)\in S\) and the one-step recursion has a form \(s_v(t+1)=f(s_v(t),v(t))\).

To prove uniqueness, we consider two different trajectories for the same input. For every $t_0<t$ both states should lie in the ball of radius $C_s$ thus 

\[
\|s_v(t)- \hat s_v(t)\|_P
\le \tau^{t-t_0}\|s_v(t_0)-\hat s_v(t_0)\|_P
\le 2C_s\tau^{t-t_0}
\]

where the right hand side will tend to zero as $t_0 \to -\infty$ therefore $s_v(t)$ should be equal to $\hat s_v(t)$ for all $t$. 

As for UEC, we compare an arbitrary trajectory started from $s_0$ at some $t_0$ with \(s_v(t_0)\) thus 
\[
\begin{aligned}
&\bigl\|s_S(s_0,t_0,v;t)-s_v(t)\bigr\|_2\\
&=\bigl\|s_S(s_0,t_0,v;t)-s_S(s_v(t_0),t_0,v;t)\bigr\|_2\\
&\le C\tau^{t-t_0}\bigl\|s_0-s_v(t_0)\bigr\|_2
\end{aligned}
\]

where $C=\sqrt{\lambda_{max}/\lambda_{min}}$ as before. 
\end{pf}

\begin{pf}[Proof of Lemma~\ref{lemma:LTI:classS_g}]
We will construct a combined system which from the innovation noise will generate the designated $w(t) =(y(t),u(t))$. As $e_g = (e_s,e_u)$ and $u(t) = G_u s_u(t)+e_u(t)$ thus 
\[
\begin{gathered}
 A_g=\begin{bmatrix}A_0&B_0G_u\\0&F_u\end{bmatrix},
 \qquad K_g=\begin{bmatrix}K_0&B_0\\0&H_u\end{bmatrix},\\
 C_g=\begin{bmatrix}C_0&0\\0&G_u\end{bmatrix}.
\end{gathered}
\]

Since $A_g$ upper triangular and both $A_0$ and $F_u$ are Schur $A_g$ is also Schur. Let $\lambda(P)$ the spectral radius of $P$. By choosing $\lambda(A_g)< \tau_g <1$ we construct a norm with $P\succ0$ and $0<\tau_g<1$ from $A_g/\tau_g$. Note, $\lambda(A_g/\tau_g) <1$. Now we define $P = \sum_{i=0}^{\infty} ((A_g/\tau_g)^i)^T(A_g/\tau_g)^i$ therefore $A_g^\top P A_g\preceq \tau_g^2P$. 

Next, we focus on innovation $e_g(t)$. It is bounded thus we may define the compact set $\mathbb V_g=\{v:\|v\|_2\le E\}$ where for all $t$, $\|e_g(t)\|_2 \leq E < \infty$ for almost surely. Since $\|A_g s\|_P \leq \tau_g \|s\|_P$ and for every $v \in \mathbb V$, $\|K_gv\|_P=\|P^{1/2}K_gv\|_2 \le \|P^{1/2}K_g\|_2 E$ the state 

\[
\begin{aligned}
\|A_gs+K_gv\|_P
&\le \|A_gs\|_P+\|K_gv\|_P\\
&\le \tau_g \|s\|_P+  \|P^{1/2}K_g\|_2 E.
\end{aligned}
\]

Based on this we may construct $\mathbb S_g=\{s:\|s\|_P\le R_g = \frac{\|P^{1/2}K_g\|_2 E}{1-\tau_g}\}$ and $f_g(s,v) = A_gs+K_gv \in\mathbb S_g$. Finally, the output $g_g(s,v)=C_gs+v$ is also bounded as 

\[
\|C_gs+v\|_2
\le \|C_gP^{-1/2}\|_2\,\|s\|_P+\|v\|_2
\le \|C_gP^{-1/2}\|_2R_g+E.
\]

therefore $\mathbb Y_g = \{w: \|w\|_2 \leq C_gP^{-1/2}\|_2R_g+E \}$. 

The Jacobian of \(f_g\) with respect to \(s\) is \(A_g\) at every \((s,v)\) thus $J_s f_g(s,v)^\top P J_s f_g(s,v)  = A_g^\top P A_g \preceq \tau_g^2P$ showing contraction. 

If we fix the state $s$ the first part of $f_g$ will cancel out and 
\[
\|f_g(s,v_1)-f_g(s,v_2)\|_2
=\|K_g(v_1-v_2)\|_2
\le \|K_g\|_2\,\|v_1-v_2\|_2
\]
 proving the lipschitzness in input. Finally, by the triangle inequality 
   \[
   \|g_g(s_1,v_1)-g_g(s_2,v_2)\|_2
   \le \|C_g\|_2\,\|s_1-s_2\|_2+\|v_1-v_2\|_2.
   \]
   
Both maps also satisfy $f_g(0,0)=g_g(0,0)=0$.

Starting from $s_0\in\mathbb S_g$ at time $t_0$ we iterate \(s(t+1)=A_gs(t)+K_ge_g(t)\) to obtain
\[
s_{\mathcal S_g}(s_0,t_0,e_g;t)
=A_g^{t-t_0}s_0+
\sum_{j=0}^{t-t_0-1}A_g^jK_ge_g(t-1-j).
\]

Now, we fix $t$ and let $t_0\to-\infty$ so the first part tends to zero and each element in the sum of the second term converges absolutely by contraction, has a \(P\)-norm at most \(\tau_g^j \|P^{1/2}K_g\|_2 E\). Stationarity follows if we take the sum to the infinite past 
\[
s_g(t)=\sum_{j=0}^{\infty} A_g^jK_g e_g(t-1-j).
\]

By \ref{eq:predictor_lti_input}, the stationary state $s_g$ has output $C_gs_g(t)+e_g(t)=w(t)$.
\end{pf}

\begin{pf}[Proof of Lemma~\ref{classS:LTI:predictor}]
We will follow steps from the proofs of Lemma~\ref{lemma:P:classS} and Lemma~\ref{lemma:LTI:classS_g}. 
We will use $w(t)$ as the input to the predictor
\[
\begin{aligned}
&\hat s(t+1)=\hat A_\theta\hat s(t)+\hat B_\theta w(t) \\
&\hat y(t)=\hat C_\theta\hat s(t)+\hat D_\theta w(t) \\
&\hat s(0)=0.
\end{aligned}
\]

From Lemma~\ref{lemma:LTI:classS_g} $\mathbb W=\mathbb Y_g$ and let us define
\[
W=\sup_{w\in\mathbb W}\|w\|_2,
\qquad
B^*=\sup_{\theta\in\Theta}\|\hat B_\theta\|_2.
\]
Since both are finite and the uniform contraction bound $\sup_{\theta\in\Theta}\|\hat A_\theta\|_2\le\tau<1$ we may take $P=I$.
Let $\|\hat s(t)\|_2\le R$ then for every $w(t)$ and $\theta$,
\[
\begin{aligned}
\|\hat s(t+1)\|_2
&\le \|\hat A_\theta\|_2\,\|\hat s(t)\|_2
   +\|\hat B_\theta\|_2\,\|w(t)\|_2\\
&\le \tau R+B^*W.
\end{aligned}
\]

Now, we choose $R=B_*W/(1-\tau)$ and $\tau R+B^*W=R$ therefore $\mathbb S=\{s:\|s\|_2\le R\}$ which also contains the initial state $0$ and every subsequent state in the predictor's trajectory. 

As of Assumption 4.2 $v=(w,\theta)$ is the input and therefore compare two updates in which the state, observation, and parameter can all differ. 

Next we consider $\Delta f = \hat f(s_1,w_1,\theta_1)-\hat f(s_2,w_2,\theta_2)$ and let us denote the Lipschitz constants of $\hat A_{\theta}$ and $\hat B_{\theta}$ by $L_A$ and $L_B$ respectively. 
Similarly to the proof in Lemma~\ref{lemma:P:classS}, we have

\[
\begin{aligned}
\Delta f
={}&\hat A_{\theta_1}(s_1-s_2)
  +\hat B_{\theta_1}(w_1-w_2)\\
 &+(\hat A_{\theta_1}-\hat A_{\theta_2})s_2
  +(\hat B_{\theta_1}-\hat B_{\theta_2})w_2
\end{aligned}
\]

and since $\| \hat A_{\theta_1}(s_1-s_2) \|_2 \leq \tau \|s_1-s_2\|_2$, $\|\hat B_{\theta_1}(w_1-w_2)\|_2 \leq B^*\|w_1-w_2\|_2$, 
$\|(\hat A_{\theta_1}-\hat A_{\theta_2})s_2 \|_2 \leq L_A R \|\theta_1-\theta_2\|_2$ and $\|(\hat B_{\theta_1}-\hat B_{\theta_2})w_2 \|_2 \leq L_B W \|\theta_1-\theta_2\|_2$ thus

\[
\begin{aligned}
\|\Delta f\|_2
&\le \tau\|s_1-s_2\|_2
+B^*\|w_1-w_2\|_2\\
&+(L_A R+L_B W)\|\theta_1-\theta_2\|_2.
\end{aligned}
\]

Now to form the inequality for $v = (w,\theta)$ and by $
\|v_1-v_2\|_2^2=\|w_1-w_2\|_2^2+\|\theta_1-\theta_2\|_2^2$ which we can put into the previous inequality to get 
$\|\Delta f\|_2
\le \tau\|s_1-s_2\|_2+L_f\|v_1-v_2\|_2$ 
where $L_f=\sqrt{(B^*)^2+(L_A R+L_B W)^2}$.

Now, we have $\hat g(s,w,\theta)=\hat C_\theta s+\hat D_\theta w$. Let us denote the Lipschitz constants of $\hat C_{\theta}$ and $\hat D_{\theta}$ by $L_C$ and $L_D$ respectively  and define $C^*=\sup_{\theta\in\Theta}\|\hat C_\theta\|_2$ and $D^*=\sup_{\theta\in\Theta}\|\hat D_\theta\|_2$. Since $\|s\|_2\le R$ and $\|w\|_2\le W$ we have 
\[
\|\hat g(s,w,\theta)\|_2 \le C^*R+D^*W
\]
which yields us to $\mathbb Y$. As before 
\[
\begin{aligned}
\Delta g&=\hat g(s_1,w_1,\theta_1)-\hat g(s_2,w_2,\theta_2)\\
&=\hat C_{\theta_1}(s_1-s_2)+\hat D_{\theta_1}(w_1-w_2)\\
&+(\hat C_{\theta_1}-\hat C_{\theta_2})s_2
 +(\hat D_{\theta_1}-\hat D_{\theta_2})w_2.
\end{aligned}
\]

Similarly to the argument about $\Delta f$ the norms are bounded as 
 $L_C R\|\theta_1-\theta_2\|_2$ and $L_D W\|\theta_1-\theta_2\|_2$. Hence
\[
\begin{aligned}
\|\Delta g\|_2\le& C^*\|s_1-s_2\|_2+D^*\|w_1-w_2\|_2\\
&+(L_C R+L_D W)\|\theta_1-\theta_2\|_2
\end{aligned}
\]
and finally with $v_i=(w_i,\theta_i)$ we may combine the previous inequalities as 
\[
\|\Delta g\|_2
\le \bar L_{g,1}\|s_1-s_2\|_2+\bar L_{g,2}\|v_1-v_2\|_2
\]
where $L_{g,1}=C_*$ and $L_{g,2}=\sqrt{D_*^2+(L_C R+L_D W)^2}$. 

Next we show UEIIS and UEC. 

For UEIIS we consider two trajectories $s_1,s_2$ initialized at $s_{0,1},s_{0,2}\in\mathbb S$ at time $t_0$ and the inputs $v_1,v_2$ thus by iteration we have  
\[
\begin{aligned}
\|s_1(t)-s_2(t)\|_2
&\le \tau^{t-t_0}\|s_{0,1}-s_{0,2}\|_2\\
&\quad+L_f\sum_{k=t_0}^{t-1}
\tau^{t-1-k}\|v_1(k)-v_2(k)\|_2.
\end{aligned}
\]

Now, let us fix $v$ and $t$, and write $s^{(t_a)}(t)$ for the state initialized at zero at time $a$. For $t_a<t_b<t$, invariance and contraction give $\|s^{(t_a)}(t)-s^{(t_b)}(t)\|_2\le R\tau^{t-t_b}$. Hence $s^{(t_a)}(t)$ has a limit $s_v(t)\in S$ as $t_a \to -\infty$. By continuity we have state recursion. Any two entire $S$-valued solutions differ at time $t$ by at most $2R\tau^{t-t_b}$ for every $t_b<t$. 

For UEC set $v_1=v_2=v$ and $s_{0,2}=s_v(t_0)$ which will lead to 
\[
\bigl\|s_{\mathcal S_{param}}(s_0,t_0,v;t)-s_v(t)\bigr\|_2
\le \tau^{t-t_0}\|s_0-s_v(t_0)\|_2.
\]
\end{pf}

\begin{pf}[Proof of Lemma \ref{ge:systemS:bilinear}] 
Finally, \(\hat s_s(\theta)=0\) is Lipschitz, and holding \(\theta\) constant recovers the claimed inequality. 
  Define $\mathbb{V}=\{ \vv= \begin{bmatrix} \vu^T & \ve \end{bmatrix}^T \in \mathbb{R}^{n_u+n_y} \mid \|\vu\|_{\infty} \le 1, \|\ve\|_{\infty} \le 1 \}$
  Let $P_g > 0$ such that $\sum_{i=0}^{n_u} A_i^T P_g A_i- \tau^2/(1+n_u) P_g \prec 0$. By \cite{costa2006discrete} such a $P_g > 0$
  always exists. Let $\vv=\begin{bmatrix} \vu^T & \ve \end{bmatrix}^T \in \mathbb{R}^{n_u+n_y}$ and let
  $f_g(\vs,\vv)=(A_0+\sum_{i=1}^{n_u}A_i\vu_i)\vs+(K_0+\sum_{i=1}^{n_u} K_i\vu_i)\ve$, 
  $g_g(\vs,\vv)=C_0\vs+\ve$. It then follows $\dfrac{f_g}{d\vs}^T(\vs,\vv) P_g \dfrac{f_g(\vs,\vv)}{d\vs} \preceq P_g$
  for all $\vv \in \mathbb{V}$.
  Indeed, by using $\vu_i\vu_j A_i^TP_gA_j \preceq \frac{1}{2}(\vu_i^2 A_i^TP_gA_i + \vu_j^2 A_j^TP_g A_j)$ for $i,j \in \{0,\ldots,n_u\}$
  with $\vu_0=1$ and using that $\vu_i^2 \le 1$. It then follows that
  $\dfrac{f_g(\vs,\vv)}{d\vs}=(A_0+\sum_{i=1}^{n_u} A_i \rvu_i)$ and hence
 $\dfrac{f_g(\vs,\vv)}{d\vs}^T P_g \dfrac{f_g(\vs,\vv)}{d\vs}=\sum_{i,j=0}^{n_u} \vu_i \vu_j A_i^T P_g A_j \le (1+n_u)\sum_{i=0}^{n_u} A_i^T P_g A_i \preceq \tau^2 P_g$. 
 From \cite[Theorem 1]{PavlovTac2011} it then follows that for 
 $\mathbb{S}=\{vs \mid \|\vs\|_2 \le \frac{\sqrt{\|P_g\|_2}}{1-\tau_g} (\sum_{i=0}^{n_u} \|K_i\|_2) n_y \|e^s(0)\|_{\infty} \}$
 and $\mathbb{Y}=\{\vy \mid \|vy\|_2 \le (\|C_0\|_2 \frac{\sqrt{\|P_g\|_2}}{1-\tau_g} \sum_{i=0}^{n_u} \|K_i\|_2) +1) n_y \|e^s(0)\|_{\infty}  \}$
 $\mathcal{S}_g=(\mathbb{S},\mathbb{V},\mathbb{Y},f_g,h_g)$ is a system of class \class{}
 with $L_g=\sqrt{\|P_g\|_2}(\sum_{i=0}^{n_u} \|K_i\|_2)$, $C=\sqrt{\|P_g\|_2}$,$\tau=\tau_g$, $L_{g,1}=\|C_0\|_2$, $L_{g,2}=1$. 
 Moreover, it is easy to see that $\rvy_{\mathcal{S}_g}(\rve_g) = \rvy(t)$ for $\rve_g=\begin{bmatrix} (\e^{s})^T & \rvu^T \end{bmatrix}^T$.
 \end{pf}
 
 \begin{pf}[Proof of Lemma \ref{ge:classS:bilinear:predictor}]
 We consider \(v=(y,u,\theta)\) and \(f(s,v)=\hat A(\theta,u)s+\hat K(\theta,u)y\). Since \(u_0=1\) and \(|u_i|\le1\) we have
\[
\|\hat A(\theta,u)x\|_P^2
\le (1+n_u)\sum_{i=0}^{n_u}\|\hat A_i(\theta)x\|_P^2
\le \hat\tau^2\|x\|_P^2.
\]
Let $b=\sup_v\|\hat K(\theta,u)y\|_P$ finite and bounded. We set $R=b/(1-\hat\tau)$ and since $\|f(s,v)\|_P\le\hat\tau R+b=R$ the set $\mathbb S=\{s:\|s\|_P\le R\}$ is invariant. 

Next, let us focus on \(g(s,v)=\hat C(\theta)s\) which is also bounded on this set. On the bounded state and input sets, the coefficient maps are bounded and Lipschitz in \(\theta\), and \(\hat A(\theta,u)\) and \(\hat K(\theta,u)\) are affine in \(u\). Hence \(f(s,\cdot)\) is Lipschitz in \(v=(y,u,\theta)\) uniformly over \(s\in\mathbb S\), with some finite constant \(L_f\). Consequently,
\[
\|f(s_1,v_1)-f(s_2,v_2)\|_P
\le\hat\tau\|s_1-s_2\|_P+L_f\|v_1-v_2\|_2.
\]

Since $\hat C$ is bounded and Lipschitz on $\Theta$ and $\mathbb S$ is bounded, $g(s,v)=\hat C(\theta)s$ is Lipschitz in $s$ and $v$. Through iteration we have UEIIS, and with a similar argument as in Lemma~\ref{classS:LTI:predictor} we will have UEC. Taking $\hat s_s(\theta)=0$ and by fixing $\theta$ we recover the predictor \eqref{pred:bilinear}.

%Iterating the inequality yeilds to UEIIS. The same argument as in the proof of Lemma~\ref{classS:LTI:predictor} will result in UEC.
% Let us define $f_{\theta}(\vs,\vx)=(A_{0}(\theta)+\sum_{i=1}^{n_u} \vu_i A_{i}(\theta))\vs+(K_{0}(\theta)+\sum_{i=1}^{n_u} K_{i}(\theta))\vy$
% and $g_{\theta}(\vs,\vx)=\hat{C}_{0}(\theta)\vs$, where $\vx=\begin{bmatrix} \vy^T & \vu^T \end{bmatrix}^T$.
% It then follows that there exists $P > 0$ such that $\sum_{i=0}^{n_u} \hat{A}_{i,\theta}^T P \hat{A}_{i,\theta} - \hat{\tau}^2/(1+n_u) P \prec 0$ and by repeating the argument showing that $\mathcal{S}_g$ is a class \class{} it follows that
% $\dfrac{f_{\theta}(\vs,\vx)}{d\vs}^T P \dfrac{f_{\theta}(\vs,\vx)}{d\vs} \preceq \hat{\tau}^2 P$.
% for all $\vx \in \mathbb{V}$. 
% Set $\hat{C}=\sqrt{\|P\|_2}$. 
% Hence, by setting $\hat{\mathbb{S}}=\{ \vs \mid \|\vs\|_2 \le \frac{\hat{C}\sqrt{n_u+n_y}\hat{L}B^{\rvq}}{1-\hat{\tau}} \}$, 
% $\hat{\mathbb{V}}=\{\vv \mid \|\vv\|_{\infty} \le B^{\rvq}\}$, 
% $\hat{\mathbb{Y}}=\{\vy \mid \|\vy\|_2 \le \frac{\hat{L}_{g,1}\sqrt{n_u+n_y}\hat{L}B^{\rvq}}{1-\hat{\tau}} \}$
% and defining $\mathcal{S}_{\theta}$ as in Assumption \ref{as:parametrisation}, it follows that
% $\mathcal{S}_{\theta}$ is a system of class \class{} and 
% $h_{\theta}(\{\rvw(s)\}_{s=0}^{t}) = \rvy_{\mathcal{S}_{\theta}}(s_0,0,\rvw;t)$.
\end{pf}
 \begin{pf}[Proof of Lemma \ref{lemma:bilinear:markov1}]
Let \(I_K=\{w\in\Sigma^+:|w|\le K\}\) and \(\Delta M(w)=M(w)-M_\theta(w)\). Stack \(z_w^y(t)\), \(w\in I_K\), into a column \(Z_K(t)\), and concatenate the matrices \(\Delta M(w)\) in the same order into a block row \(\Delta M_K\). Write \(T_K^y=\mathbb E[Z_K(t)Z_K(t)^\top]\). For every sufficiently large \(K\), the excitation assumption gives
\[
\begin{aligned}
\mathbb E\|\Delta M_K Z_K(t)\|_2^2
&=\operatorname{tr}\!\left(\Delta M_K T_K^y\Delta M_K^\top\right)\\
&\ge m_y\operatorname{tr}\!\left(\Delta M_K\Delta M_K^\top\right)\\
&=m_y\sum_{w\in I_K}\|\Delta M(w)\|_F^2.
\end{aligned}
\]
Unrolling the true predictor and the fitted predictor for \(K\) steps gives
\[
\Delta M_K Z_K(t)\longrightarrow y(t)-e^s(t)-\hat y_\theta(t)
\quad\text{in }L^2.
\]
Indeed, each omitted output term is a bounded readout applied to a product of \(K\) contracting state matrices and a bounded past state. Its norm therefore decays geometrically. Since \(Z_K(t)\) depends only on observations through \(t-1\), the centered current innovation \(e^s(t)\) is orthogonal to \(\Delta M_K Z_K(t)\), and hence to its \(L^2\) limit. Consequently,
\[
L(\theta)
=\sigma_{e^s}^2+
\lim_{K\to\infty}\operatorname{tr}\!\left(\Delta M_K T_K^y\Delta M_K^\top\right)
\ge \sigma_{e^s}^2+m_y\|M-M_\theta\|_{H_2}^2.
\]

% \begin{equation}
% \begin{aligned}
%     &\mathcal{L}(\theta)=\sigma^2_{\e^s} +\sum_{k,l=0}^{\infty} \sum_{q_1,q_2 \in \Sigma, w,v \in \Sigma^{*},|w|=k}
%    \mathrm{tr} (\hat{C}(\theta_{true}) \hat{A}(\theta_{true})_{w} \\
%    &\hat{K}_{q_1}(\theta_{true}) - \hat{C}(\theta) (A_{\theta})_w (\hat{K}_\theta)_{q_1}) E[\rvz^{\rvy}_{wq_1}(t)\rvz^{\rvy}_{vq_2}(t)^T](\hat{C}(\theta_{true}) \\
%    &\hat{A}(\theta_{true})_{v} \hat{K}_{q_2}(\theta_{true}) - \hat{C}(\theta) (A_{\theta})_v (\hat{K}_\theta)_{q_2})^T  
% \end{aligned}
% \end{equation}
% here $\sigma^{2}_{\e^s}=E[\|\e^s(t)\|^2_2]$.
% Note that by \cite{TAC:MPRV}, $E[\rvz^{\rvy}_{q_1w}(t)(\rvz_{q_2v}^{\rvy}(t))^T]=0$ if $q_1w \ne q_2w$ and 
% $E[\rvz_{qw}^{\rvy}(t)(\rvz_{qw}^{\rvy}(t))^T]=E[\rvz_q(t)(\rvz_q(t))^T] \triangleq \Lambda_q$
% \begin{equation}
% \begin{aligned}
%  \mathcal{L}(\theta)&=\sigma^2_{\e^s}+\sum_{k=0}^{\infty} \sum_{q \in \Sigma, w \in \Sigma^{*},|w|=k}
%    \mathrm{tr} (\hat{C}(\theta_{true}) \hat{A}(\theta_{true})_{w} \\
%    &\hat{K}_{q}(\theta_{true}) - \hat{C}(\theta) (A_{\theta})_w (\hat{K}_\theta)_{q}) 
%    \Lambda_q (\hat{C}(\theta_{true}) \hat{A}(\theta_{true})_{w} \\
%    &\hat{K}_{q}(\theta_{true}) 
%    - \hat{C}(\theta) (A_{\theta})_w (\hat{K}_\theta)_{q})^T
% \end{aligned}
% \end{equation}
\end{pf}
 \textbf{Existence of a wide-sense stationary solution of \eqref{eq:T0sys_bilinear}}
We will use the following notation: let $\Sigma=\{0,\ldots,n_u\}$ and denote by $\Sigma^{*}$ the set of all finite sequences over $\Sigma$;
  we denote by $\epsilon$ the empty sequence; we denote by $|w|$ the length of a sequence $w \in \Sigma^*$, by definition $| \epsilon | = 0$. For a collection of square matrices $\{A_i\}_{i \in \Sigma}$ and a sequence $\sigma=\sigma_1\ldots\sigma_k \in \Sigma^*$, we denote $A_{\sigma}=A_{\sigma_k}\cdots A_{\sigma_1}$ and $A_{\epsilon}=I$.
  For any process $\rvr$, define $\rvz_w^{\rvr}(t)$ recursively as follows: $\rvz^{\rvr}_{\epsilon}(t)=\rvr(t)$ and $\rvz^{\rvr}_{wq}(t)=\rvz^{\rvr}_{w}(t-1)\rvu_q(t-1)$ for $w \in \Sigma^*, q \in \Sigma$.

  From \cite{TAC:MPRV} it follows that 
  \begin{equation}
    \label{eq:bilinear_output}
    \begin{aligned}
      \rvy(t)&=\e^s(t)+\sum_{k=0}^{\infty} \sum_{q\in \Sigma, w \in \Sigma^*, |w|=k} C_0 A_w K_q \rvz^{\e^s}_{wq}(t)
      \rvs(t)\\
      & = \sum_{k=0}^{\infty} \sum_{q \in \Sigma, w \in \Sigma^*, |w|=k} A_wK_q \rvz^{\e^s}_{wq}(t-1)
    \end{aligned}
  \end{equation}

\end{document}